\documentclass[format=acmsmall, screen=true, review=false,authorversion,nonacm]{acmart}

\AtBeginDocument{%
    \providecommand\BibTeX{{%
            \normalfont B\kern-0.5em{\scshape i\kern-0.25em b}\kern-0.8em\TeX}}}

\acmYear{2021}
\acmDOI{10.1145/3480172}

\acmJournal{TIOT}

\usepackage{textcomp}
\usepackage{xcolor}
\usepackage[linesnumbered,ruled]{algorithm2e}

\usepackage{mathrsfs}
\usepackage{multirow}
\usepackage{bm}

\usepackage{epstopdf}
\usepackage[caption=true,font=footnotesize]{subfig}

\newcommand{\STAB}[1]{\begin{tabular}{@{}c@{}}#1\end{tabular}}

\definecolor{navyblue}{rgb}{0.0, 0.0, 1.0}
\newcommand{\navy}[1]{{\color{navyblue}{#1}}}
\newcommand{\at}{\makeatletter @\makeatother}

\setcopyright{none}

\begin{document}

\title{Adaptive Anomaly Detection for Internet of Things in Hierarchical Edge Computing: A Contextual-Bandit Approach}

\author{Mao~V.~Ngo}
\affiliation{%
    \institution{Singapore University of Technology and Design}
    \streetaddress{8 Somapah Road}
    \postcode{487372}
    \country{Singapore}
}
\email{vanmao\_ngo@mymail.sutd.edu.sg}

\author{Tie Luo}
\affiliation{%
    \institution{Missouri University of Science and Technology}
    \city{Rolla}
    \state{Missouri}
    \postcode{65401}
    \country{USA}
}
\email{tluo@mst.edu}
\authornote{Tie Luo is the corresponding author.}


\author{Tony Q.S. Quek}
\affiliation{%
    \institution{Singapore University of Technology and Design}
    \streetaddress{8 Somapah Road}
    \postcode{487372}
    \country{Singapore}
}
\email{tonyquek@sutd.edu.sg}

\authorsaddresses{%
	Authors' addresses: Mao V. Ngo and Tony Q.S. Quek, Singapore University of Technology and Design, 8 Somapah Road, 487372, Singapore, {vanmao\_ngo\at mymail.sutd.edu.sg}, {tonyquek\at sutd.edu.sg}; Tie Luo, Department of Computer Science, Missouri University of Science and Technology, Rolla, MO 65401, USA, {tluo\at mst.edu}.
}


\begin{abstract}
\iffalse 
The advances in deep neural networks (DNN) have significantly enhanced real-time detection of anomalous data in Internet-of-Things (IoT) applications.
However, the complexity-accuracy-delay dilemma persists: while complex DNN models can offer higher accuracy, typical IoT devices can barely afford the computation load, and the remedy of offloading the load to the cloud tends to incur long delay. In this paper, we address this challenge in the context of anomaly detection in IoT, by proposing an adaptive anomaly detection scheme with hierarchical edge computing (HEC).
Specifically, we first construct multiple anomaly detection DNN models with increasing complexity, and associate each of them to a corresponding layer in an HEC architecture. 
Then, we design an adaptive model selection scheme to automatically select the best suited one among these models. We formulate this model selection problem as a {\em contextual-bandit problem} characterized by a single-step Markov Decision Process, and solve it using a {\em reinforcement learning policy network}.
Next, we build an HEC testbed using real IoT devices, and implement our proposed contextual-bandit approach on the testbed. In particular, we also incorporate an accelerated policy training method that introduces parallelism into the training process, by taking advantage of the nature of distributed models. We evaluate our approach using both univariate and multivariate IoT datasets in our HEC testbed, in comparison with both baseline and state-of-the-art schemes. We show that our adaptive approach strikes the best accuracy-delay tradeoff with the univariate dataset, and achieves the best accuracy and F1-score while the delay is negligibly larger than the best scheme with the multivariate dataset.
\else 
The advances in deep neural networks (DNN) have significantly enhanced real-time detection of anomalous data in IoT applications. However, the complexity-accuracy-delay dilemma persists: complex DNN models offer higher accuracy, but typical IoT devices can barely afford the computation load, and the remedy of offloading the load to the cloud incurs long delay. In this paper, we address this challenge by proposing an adaptive anomaly detection scheme with hierarchical edge computing (HEC). Specifically, we first construct multiple anomaly detection DNN models with increasing complexity, and associate each of them to a corresponding HEC layer. Then, we design an adaptive model selection scheme that is formulated as a {\em contextual-bandit problem} and solved by using a {\em reinforcement learning policy network}. We also incorporate a parallelism policy training method to accelerate the training process by taking advantage of distributed models. We build an HEC testbed using real IoT devices, implement and evaluate our contextual-bandit approach with both univariate and multivariate IoT datasets. In comparison with both baseline and state-of-the-art schemes, our adaptive approach strikes the best accuracy-delay tradeoff on the univariate dataset, and achieves the best accuracy and F1-score on the multivariate dataset with only negligibly longer delay than the best (but inflexible) scheme.
\fi
\end{abstract}



\maketitle

\section{Introduction}
\label{sec:introduction}

The proliferation of smart Internet of Things (IoT) devices has spurred global deployment of smart factories, smart homes, autonomous vehicles, digital health, and so on.
The gigantic network of these IoT devices generates a sheer amount of sensory data that can be exploited to extract knowledge insights or detect anomalous events, by leveraging recently developed machine learning and especially deep learning techniques~\cite{Mohammadi_SurveyIoTBigData_2018, chalapathy2019deep, Luo_ICC2018, Malhotra_LSTM_encDec_ICMLWrsh2016}.
Certain IoT applications, such as collision avoidance for autonomous vehicles, fire alarm systems in factories, or fault diagnosis of automatic industrial processes, are time-critical and require fast anomaly detection to prevent unexpected breaks that can lead to costly or even fatal failures.
In such cases, the traditional approach of streaming all the IoT sensory data to the cloud can be problematic as it tends to incur high communication delay, congest backbone network, and pose a risk on data privacy.

To this end, the emerging edge or fog computing \cite{La_FogComputing_2019, Chen_SEC2017} provides a better alternative by performing anomaly detection (AD) in the proximity of the sources of sensory data.
However, pushing computation from cloud to the edge of networks faces resource challenges especially when the model is complex (such as deep learning models) and the edge device only has limited computation power, storage, and energy supply, which is the case for typical IoT devices.

A possible solution is to transform a large complex model into one that fits the IoT device's capability.
For example, model compression \cite{Han2015DeepCompression} achieves this by pruning redundant and unimportant (near-zero) parameters as well as by quantizing weights into bins;
or Hinton \textit{et al.}~\cite{hinton2015distilling} proposed a knowledge distillation technique that transfers the knowledge learned by a large-original model to a smaller-distilled model, by training a distilled model to learn the soft output of the large model.
However, such approaches need to handle each AD model on a case-by-case basis via fine-tuning, or are only applicable to a few specific types of deep neural networks (DNNs) with large sparsity.

There are also other proposed approaches \cite{Teerapittayanon_ICDCS2017, Neurosurgeon_Kang2017} on distributed anomaly detection, but overall, we identify three main issues in most existing works:
(1) attempting ``one size fits all''---use one AD model to handle all the input data, while overlooking the fact that different data samples often have different difficulty levels in detecting anomalous events;
(2) focusing on accuracy or F1-score without giving adequate consideration to detection delay and memory footprint;
(3) lacking appropriate local analysis in distributed systems and thus often transmitting data back and forth between edge sources and the cloud, which incurs unnecessary delay and bandwidth consumption.


In this paper, we propose an adaptive distributed AD approach that leverages the hierarchical edge computing (HEC) architecture by matching input data of different difficulty levels with AD models of different complexity on the fly.
Specifically, we construct multiple anomaly detection DNN models (using autoencoder and LSTM) of increasing complexity, and associate each of them to an HEC layer from bottom to top, e.g., IoT devices, edge servers, and cloud.
Then, we propose an adaptive model selection scheme that judiciously selects one of the AD models based on the contextual information extracted online from input data.
We formulate this model selection problem as a {\em contextual bandit problem}, which is characterized by a single-step Markov Decision Process (MDP), and solve it using a {\em reinforcement learning policy network}.
The single-step MDP enables quick decision-making on model selection, and the decisions thus made avoid unnecessary data transmission between edge and cloud, while retaining the best possible detection accuracy.

We build an HEC testbed using real IoT devices, and implement our proposed contextual-bandit approach on the testbed. We evaluate our approach using both univariate and multivariate IoT datasets in our HEC testbed, in comparison with both baseline and state-of-the-art schemes. In summary, this paper makes the following contributions:



\begin{itemize}
    \item We identify three main issues in existing IoT anomaly detection approaches, namely using one universal model to fit all scenarios, one-sided focus on accuracy, and lack of local analysis that results in unnecessary network traffic.

    \item We propose an adaptive AD approach that differentiates inputs by matching data of different difficulty levels with AD models of different complexity in an HEC. Our approach uses a contextual-bandit theoretical framework and the solution is obtained via a {\em reinforcement learning policy network}.

    \item In our implementation of the proposed approach on an HEC testbed, we propose and incorporate an accelerated policy training method, which introduces parallelism by leveraging the distributed environment and achieves 4-5 times faster training time than traditional, sequential training.

    \item We build a real HEC testbed and evaluate our proposed approach with baseline and state-of-the-art schemes, using real-world IoT datasets (including both univariate and multivariate). Our approach strikes the best accuracy-delay tradeoff on the univariate dataset, and achieves the best accuracy and F1-score on the multivariate dataset with only negligibly longer delay than the best (but inflexible) scheme.

\end{itemize}

\section{Related work}
\label{sec:relatedWork}

Anomaly detection has been extensively studied in several surveys \cite{Mohammadi_SurveyIoTBigData_2018, GuptaSurveyTKDE2013, chalapathy2019deep}.
Here we give a brief discussion of some works related to our approach, and refer readers to these surveys for a more in-depth discussion.
Deep learning is becoming increasingly popular in anomaly detection for IoT applications \cite{Mohammadi_SurveyIoTBigData_2018,Luo_ICC2018,singh2017anomaly,Malhotra_LSTM_encDec_ICMLWrsh2016, su2019robust, chalapathy2019deep}.
For univariate data, Luo and Nagarajan \cite{Luo_ICC2018} proposed an autoencoder (AE) neural network-based model to learn patterns of normal data and detect outliers (based on reconstruction errors) in a distributed fashion in IoT systems.
Their AE model can be deployed at IoT devices to perform detection, but the model is fairly lightweight and may not be able to detect some complex anomalous events or high dimensional data.
For multivariate data, an LSTM-based encoder-decoder model was proposed by \cite{Malhotra_LSTM_encDec_ICMLWrsh2016} to learn normal data pattern and predict a few future time-steps based on a few historical time-steps.
Or, Su \textit{et al.} \cite{su2019robust} proposed a complex AD model, which glues GRU (a variant of RNN), variational autoencoder, and planar normalizing flows, to robustly learn temporal dependence and stochasticity of multivariate time series.
However, this model does not suit resource-constrained IoT devices due to its high computational cost.

On another line of research, some other distributed machine learning models split complex models and deploy partial models at different layers of HEC~\navy{\cite{Teerapittayanon_ICDCS2017, Neurosurgeon_Kang2017, Zhou_AAIoT2019, Li_IEEENetwork2018}}.
Zhou \textit{et al.} \cite{Zhou_AAIoT2019} presented a method to reduce the inference time of a DNN model in HEC by splitting and allocating an appropriate partial DNN model to each layer of HEC.
Teerapittayanon \textit{et al.} \cite{Teerapittayanon_ICPR2016} proposed a BranchyNet architecture for an image classification task that can early ``exit'' from a multi-layer DNN during inference based on the confidence of inference output.
Later on, the same authors \cite{Teerapittayanon_ICDCS2017} deployed different sections of BranchyNet in an HEC system, in order to reuse extracted features from lower layers to do inference at a higher layer.
This requires less communication bandwidth and allows for faster and localized inference due to the shallow model at the edge.
However, this approach has to make inferences sequentially from the very bottom to the top of HEC, which can lead to unnecessary delay and inference requests to lower layers when detection is hard.
Also, it requires all the distributed models to use the same architecture, while our approach has the flexibility of using different models at different layers of HEC.

Our work is inspired by the observation that input data often consists of different levels of difficulty in analysis (e.g., easy, medium, and complex) that should be treated differently to achieve the best inference performance. Given the generality of this idea, some similar forms appeared in prior work \cite{Taylor_SIGPLAN2018,NIPS2017_RuntimeNeuralPruning,Blockdrop_WuCVPR2018}, but there are key differences from this work.
\cite{Taylor_SIGPLAN2018} used multiple k-Nearest Neighbor (kNN) classification models to train a selection model to choose a proper inference model (among several models within an embedded device) for a given input image and desired accuracy.
Lin \textit{et al.} \cite{NIPS2017_RuntimeNeuralPruning} proposed a dynamic neural networks pruning framework at \textit{runtime} by using a reinforcement learning (RL)-based approach to decide to prune or keep each convolutional neural networks (CNN) layer conditioned on difficulty levels of input samples, resulting in reducing average inference time while achieving high accuracy overall.
Similarly,
\cite{Blockdrop_WuCVPR2018} proposed a BlockDrop scheme that learns to dynamically drop or keep a residual block of a trained deep residual networks (ResNets) during inference, to minimize the number of residual blocks and preserve recognition accuracy.
The key differences of our work from the above are as follows. First, these prior works deal with the task of image classification in the computer vision domain, while our work deals with the task of anomaly detection in IoT and edge computing. Second, using multiple sequentially kNN classifiers as in \cite{Taylor_SIGPLAN2018} does not scale well. In contrast, we use a single {\em policy network} that directly outputs a suitable model based on contextual information, without the need for checking each model (e.g., kNN) one by one until meeting a certain accuracy. It is also lightweight and runs quickly, and can be easily scaled to a large number of devices. Third, our RL-based approach deals with a distributed computing environment with multiple models, while the above prior works \cite{Blockdrop_WuCVPR2018, NIPS2017_RuntimeNeuralPruning} deal with a single model in a central environment.

This paper is a significant extension to our preliminary work \cite{Mao2020adaptive,MaoICDCS2020} including new training methods, new state-of-the-art schemes in comparison, newly designed experiments, and new results, as described in Sections \ref{subsec:Implementation}, \ref{subsec:AcclerationTrainingPolicyNetwork}, \ref{subsec:ResultAcceleration}, \ref{subsec:HandcraftedVsEncoded}. 

\section{Contextual-Bandit Anomaly Detection}
\label{sec:AdaptiveAnomalyDetection}

\subsection{Overall Approach}

\begin{figure}[tb]
    \centering
    \includegraphics[width=1.0\linewidth]{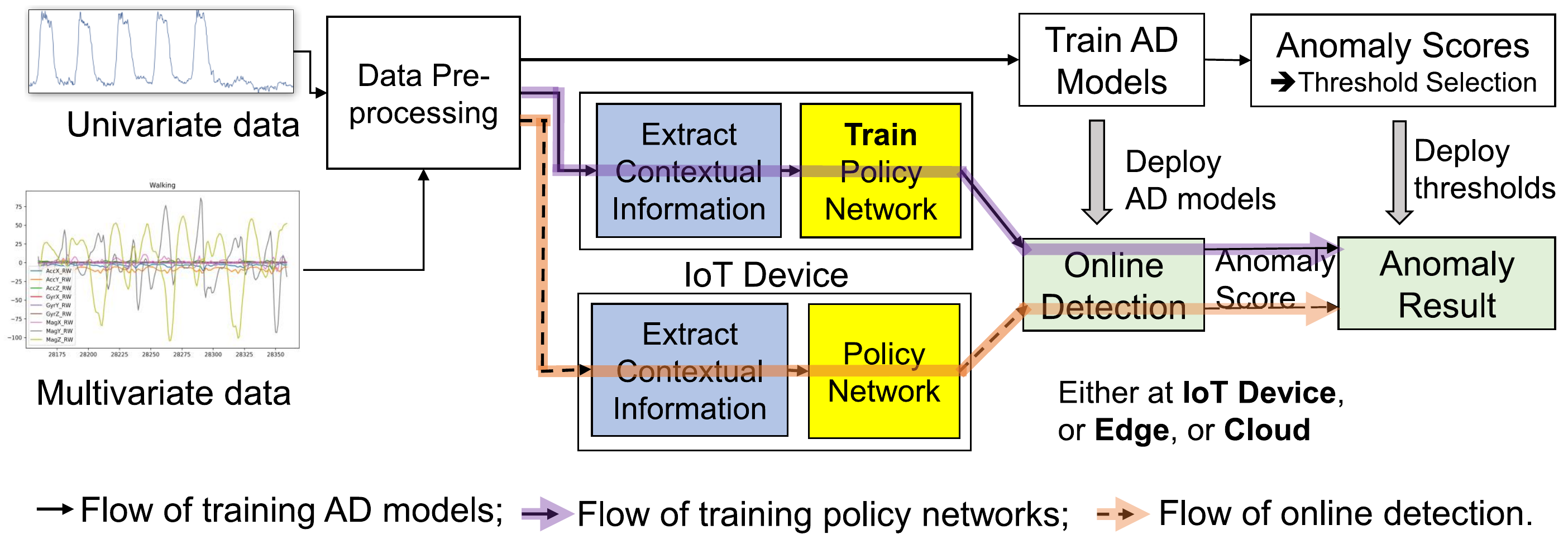}%
    \caption{Overall structure of training and online detection data flows of the anomaly detection models.
        The black solid line denotes flow of training AD models, the transparent purple line denotes flow of training policy networks, and the transparent orange line denotes flow of online detection.}
    \label{fig:DataPipeline}
\end{figure}

Fig.~\ref{fig:DataPipeline} shows the overall structure of the adaptive anomaly detection approach, which consists of three flows: (1) training multiple AD models (black solid line), (2) training policy networks (transparent purple line), and (3) online adaptive detection (transparent orange line).
A common stage for all three flows is \textit{Data Pre-processing} which has input either univariate or multivariate data.
In this stage, the training dataset is standardized to zero mean and unit variance. Then an according training scaler is used to transform the validation/test set, and used for online detection.
The standardized datasets are segmented into sequences through sliding windows.

In the first flow, \textit{Training AD Models} stage constructs multiple AD models (with increasing complexity) based on the training time-series sequences that are pre-processed.
The AD models capture normal patterns of data (either univariate or multivariate, which are presented in Subsection~\ref{subsec:multipleADModels}).
Then, the \textit{Anomaly Scores} stage finds anomaly scores and thresholds for detecting anomaly during validation and online detection phases.
These trained AD models with increasing complexity and their appropriate thresholds are deployed at different layers of HEC, from bottom to top.

In the second flow, we design and \textit{train the policy networks} based on extracted contextual information (e.g., encoded feature representation) of input sequences to adaptively select the best AD models on the fly, in order to achieve high accuracy and low detection delay simultaneously (Subsection~\ref{subsec:DynamicModelSelectionScheme}).

Finally, during the \textit{online adaptive detection} phase, the trained policy network deployed at IoT device can select the best suited AD model on the fly during testing phase.

\subsection{Constructing Multiple Anomaly Detection Models in HEC}
\label{subsec:multipleADModels}
We consider a $K$-layer distributed hierarchical edge computing (HEC) system: IoT devices at layer-1, edge servers at layer-2 to layer-$(K-1)$, and the cloud at layer-$K$.
We choose $K=3$ as a typical setting for a $K$-layer HEC \cite{La_FogComputing_2019, Mohammadi_SurveyIoTBigData_2018, Ngo_Globecom2020} (but our approach applies to any $K$ in general, i.e., multiple layers of edge servers).
We consider two types of data: univariate and multivariate.
In this subsection, we construct $K$ AD models for each data type, with increasing complexity, and associate those models with the HEC layers from 1 to $K$.

\iffalse
\begin{figure*}[tb]
    \centering
    \subfloat[AD models in HEC (with hardware choices from our testbed).]{%
        \includegraphics[width=0.66\linewidth]{Figures/OverviewSystem.eps}%
        \label{fig:OverviewSystem1}%
    }\hfill
    \subfloat[Software architecture.]{%
        \includegraphics[width=0.32\linewidth]{Figures/SoftwareArchitecture.eps}%
        \label{fig:SoftwareArchitecture1}
    }
    \caption{Overview of proposed approach: different anomaly detection (AD) models are deployed at different layers in a hierarchical edge computing (HEC) architecture, with a contextual-bandit policy network residing at each IoT device.}
\end{figure*}
\else 
\begin{figure}[tb]
    \centering
    \includegraphics[width=0.950\linewidth]{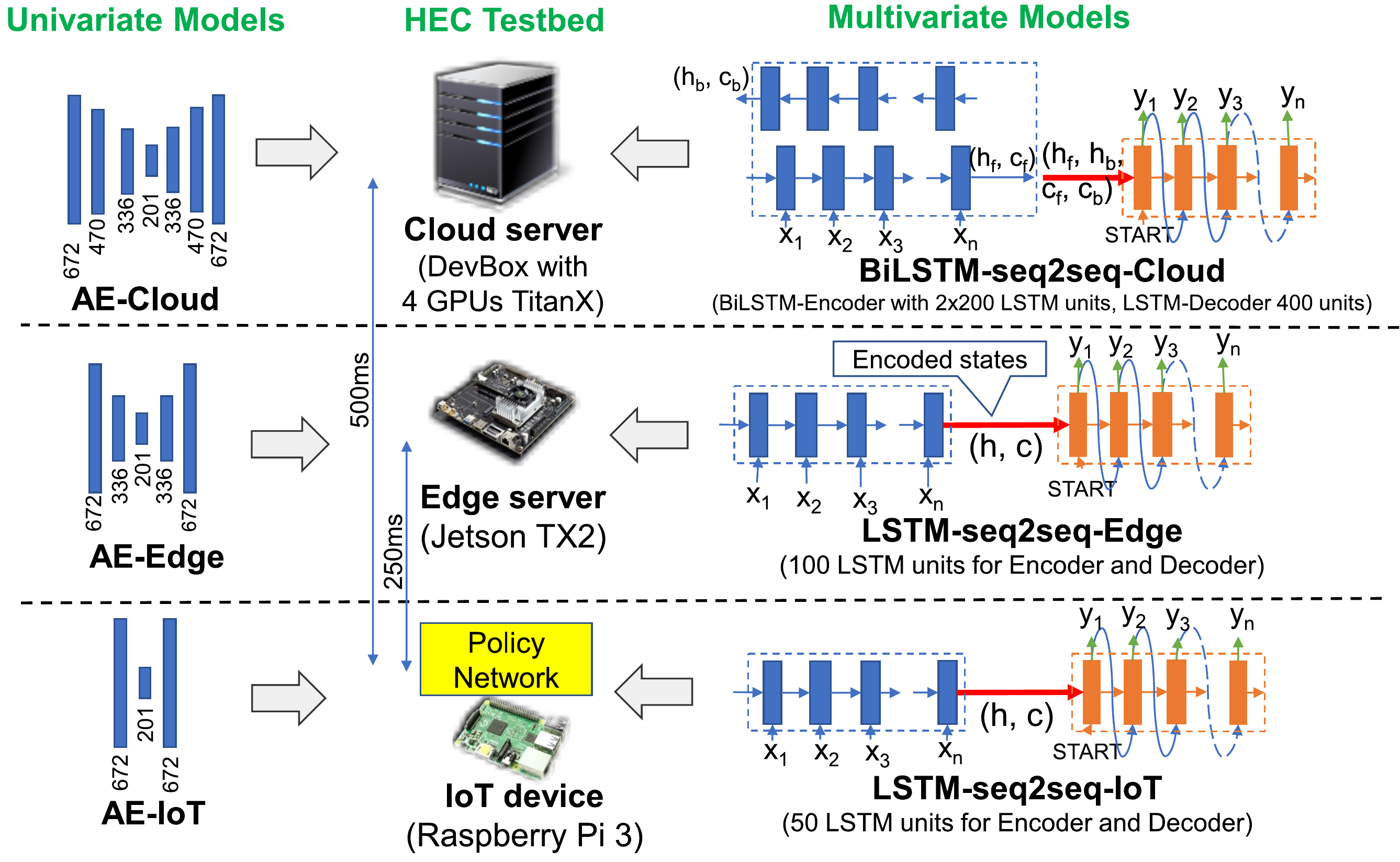}%
    \caption{Overview of proposed approach: different anomaly detection (AD) models are deployed at different layers in a hierarchical edge computing (HEC) architecture (consisting of hardware choices from our testbed), with a contextual-bandit policy network residing at each IoT device.}
    \label{fig:OverviewSystem}%
\end{figure}
\fi


\subsubsection{AD Models for Univariate Data}
\label{subsubsec:ADModelsUnivariate}

\begin{figure}[t]
    \centering
    \includegraphics[width=0.80\linewidth]{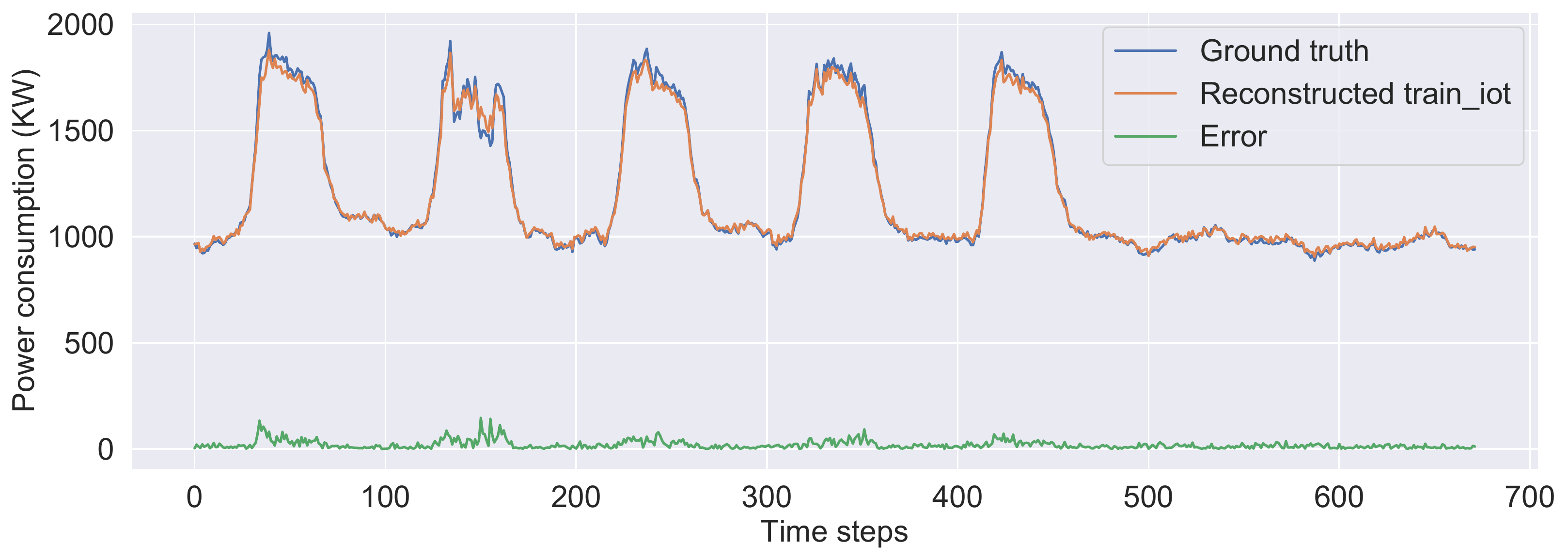}
    \caption{Reconstruction error of AE-IoT model.}
    \label{fig:ReconstructionTrainAutoencoderIoT}
\end{figure}

For univariate IoT data, we adapt the autoencoder (AE) model with a single hidden layer from \cite{Luo_ICC2018}, which has proved the feasibility of running this model on IoT devices.
In \cite{Luo_ICC2018}, the compression ratio between the dimension of the encoded layer and that of the input layer is 70\%; while in our case, we use a much lower ratio of 30\% in order to fit more diverse low-cost IoT devices. 
Simulation shows that our model under this ratio can still reconstruct normal data very well (see Fig.~\ref{fig:ReconstructionTrainAutoencoderIoT}).
For edge servers (e.g., IoT gateways or micro-servers), we consistently use the AE-based model with more hidden layers to enhance the capability of learning better features to represent data.
We add one more encoder layer and one more decoder layer to the previous AE model to obtain a model which we call \textit{AE-Edge}.
For the cloud, we further add one more encoder layer and decoder layer to have a deep AE model, which we refer to as \textit{AE-Cloud}.

The detailed setup of the above AE-based models is shown in the first column of Fig.~\ref{fig:OverviewSystem}.
Each number beside a layer of the AE-based model is the number of neural units of a corresponding layer.
We train AE-based models with stochastic gradient descent (SGD) to minimize the mean absolute reconstruction error between the reconstructed outputs and the expected outputs (which are equal to the inputs).
The reconstruction error is $e_i=|x_i-\hat{x_i}|$ where $x_i$ is input data and $\hat{x_i}$ is the corresponding reconstructed output.
To avoid overfitting, we use $\ell_2$-norm regularization for weights, and add a dropout rate of 0.3 after each hidden layer.
In accordance with the different complexities of these models, we train them over 4000, 6000 and 8000 training epochs for AE-IoT, AE-Edge, and AE-Cloud, respectively.

\subsubsection{AD Models for Multivariate Data}
\label{subsubsec:ADModelsMultivariate}
The simplicity of AE-based models does not well capture representation features for high-dimensional IoT data. Hence in the multivariate case (18 dimensions in our evaluated dataset), we use sequence-to-sequence (seq2seq) model \cite{Sutskever2014SequenceTS} based on long short-term memory (LSTM) to build an LSTM encoder-decoder as the AD model.
Such models can learn representations of multi-sensor time-series data with 1 to 12 dimensions \cite{Malhotra_LSTM_encDec_ICMLWrsh2016}.
In our case, we apply our LSTM-seq2seq model to an 18-dimensional dataset, and deploy the model on the IoT device, and name it \textit{LSTM-seq2seq-IoT}.
The multivariate IoT data are encoded into encoded states by an LSTM-encoder, then an LSTM-decoder learns to reconstruct data from previous encoded states and previous output, one step a time.
For the first step, a special token is used, which in our case is a zero vector.
For the edge layer, we build an \textit{LSTM-seq2seq-Edge} anomaly detection model with a double number of LSTM units for both encoder and decoder, which can learn a better representation of a longer sequence input.
For the cloud layer, we build a \textit{BiLSTM-seq2seq-Cloud} anomaly detection model with a bidirectional-LSTM (BiLSTM) encoder to learn both backward and forward directions of the input sequence to encode information into encoded states (i.e., by concatenating encoded states from two directions).
These are depicted in the third column of Fig.~\ref{fig:OverviewSystem}.
To train these LSTM-seq2seq models, we use a teacher forcing method \cite{BengioNIPS2015} with the \texttt{RMSProp} optimizer and $\ell_2$-norm kernel regularizer of $1e-4$ to minimize the mean squared reconstruction error.
The output of LSTM-decoder is dropped out with a drop-rate $0.3$, and then passes through a fully-connected layer with the linear activation function to generate a reconstruction sequence.

\subsubsection{Anomaly Score}

The training process shows that the above models can well capture the normal data pattern, indicated by low reconstruction error for normal data and high error for abnormal data.
Therefore, the reconstruction error is a good indicator for detecting anomalies.
We assume that reconstruction errors generally follow Gaussian distribution\footnote{We use univariate/multivariate Gaussian distribution for univariate/multivariate dataset, respectively.} $\mathcal N(\mathbf{\mu}, \mathbf{\Sigma})$, where $\mathbf\mu$ and $\mathbf{\Sigma}$ are the mean and covariance matrix of reconstruction errors of normal data samples.
We use \textit{logarithmic probability densities (logPD)} of the reconstruction errors as \textit{anomaly scores}, as in \cite{singh2017anomaly, Malhotra_LSTM_AD_ESANN2015, Mao2020adaptive}.
Normal data will have a high logPD while anomalous data will have a low logPD (note logPD is negative).
We then use the minimum value of the logPD on the normal dataset (i.e., the training set) as the threshold for detecting outliers during testing.

We consider a detection as confident if the input sequence being detected satisfies one of these two conditions:
(i) at least one data point has a logPD of less than a certain factor (e.g., 2x) of the threshold;
(ii) the number of anomalous points is higher than a certain percentage (e.g., 5\%) of the sequence size.

\subsection{Adaptive Model Selection Scheme}
\label{subsec:DynamicModelSelectionScheme}

As AD models are deployed at the IoT, edge, and cloud layers of HEC respectively, we propose an adaptive model selection scheme to select the most suitable AD model based on the contextual information of input data, so that each data sample will be directly fed to its best-suited model.
Note that this is in contrast to traditional approaches where input data will either (i) always go to one fixed model regardless of the hardness of detection \cite{Chen_SEC2017}, or (ii) be successively offloaded to higher layers until meeting a required or desired accuracy or confidence \cite{Teerapittayanon_ICDCS2017}.

Our proposed adaptive model selection scheme is a reinforcement learning algorithm that adapts its model selection strategy to maximize the expected reward of the model to be selected.
We frame the learning problem as a \textit{contextual bandit problem} \cite{sutton2000policy,williams1992REINFORCE} (which is also known as associative reinforcement learning (RL), one-step RL, associative bandits, and learning with bandit feedback) and use a single-step Markov decision process to solve it.

\begin{figure}[t]
    \centering
    \includegraphics[width=0.60\linewidth]{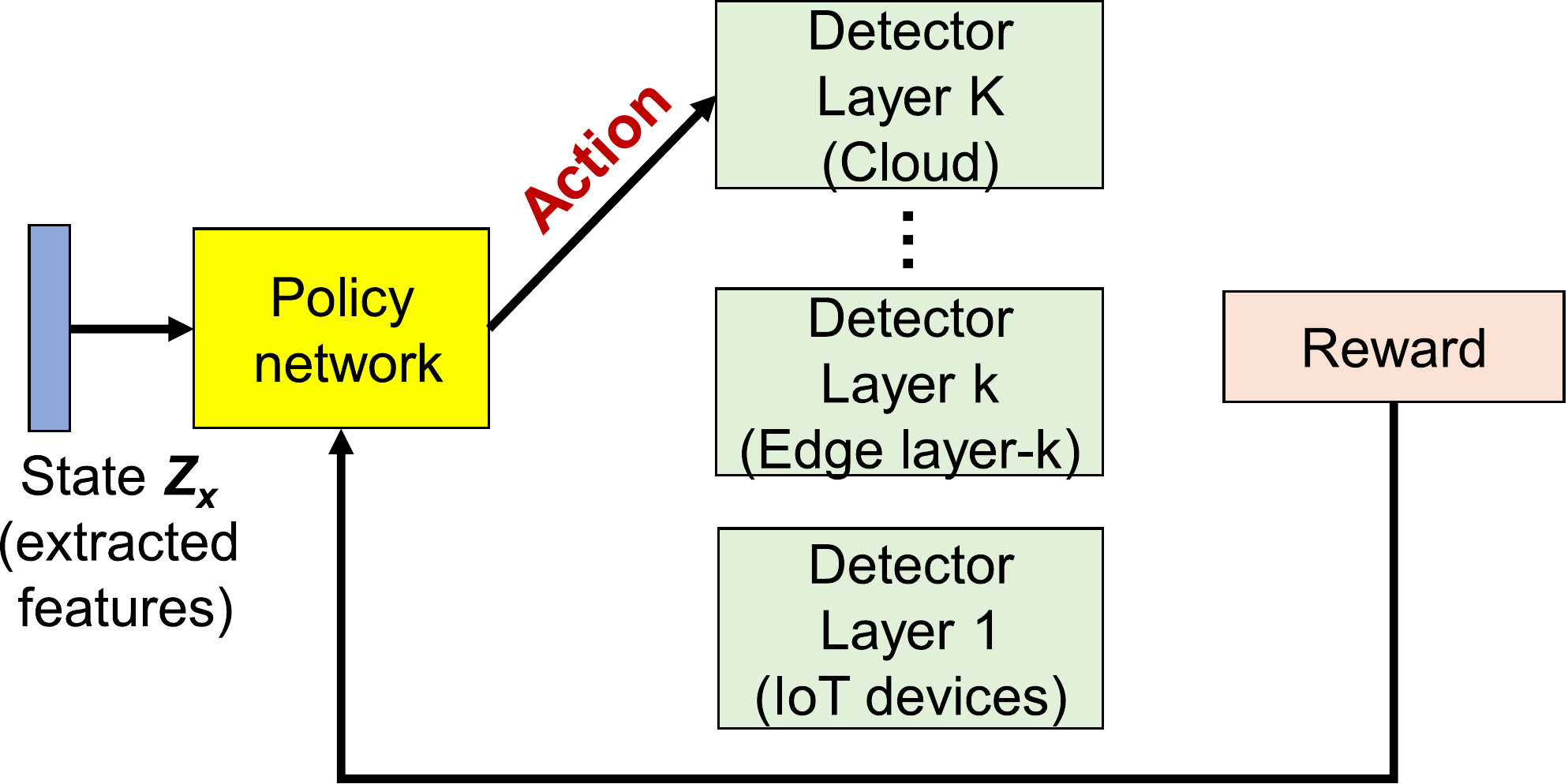}
    \Description[Decision-making module using contextual-bandit approach]{A reinforcement learning policy network provides an adaptive model selection for anomaly detection task.}
    \caption{Contextual-bandit model selection with a policy network.}
    \label{fig:DecisionMakingModule}
\end{figure}

The contextual-bandit model selection approach is illustrated in Fig.~\ref{fig:DecisionMakingModule}. Formally, given the contextual information $\mathbf{z_x}$ of an input data $\mathbf{x}$, where $\mathbf{z_x}$ is a representation of the input data, 
and $K$ trained AD models deployed at the $K$ layers of an HEC system,
we build a policy network that takes $\mathbf{z_x}$ as the input state and outputs a policy of selecting which model (or equivalently which layer of HEC) to do anomaly detection, in the form of a categorical distribution
\begin{align}
\pi_{\theta} (\mathbf{a}|\mathbf{z_x}) = \prod_{k=1}^{K} s_k^{a_k}
\label{eq:CategoricalDistribution}
\end{align}
where $\mathbf{a}=(a_1, a_2,\cdots,a_K)$, $a_k\in\{0,1\}$, is the actions encoded as a one-hot vector which defines which model (or HEC layer) to perform the task, $\mathbf{s}=(s_1, s_2,\cdots, s_K)=f_{\theta}(\mathbf{z_x})$, $s_k \in [0,1]$, is a likelihood vector representing the likelihood of selecting each model $k$, and $\sum_{k=1}^{K} a_k = 1, \sum_{k=1}^{K} s_k = 1$. We set $a_k=1$ if $k=\arg \max_{k} (s_k)$ and $a_k=0$ otherwise, and we denote the selected action as $|\mathbf{a}|=k$.

The policy network $f_{\theta}(.)$ is designed as a neural network with parameters $\theta$.
To make the policy network small enough to run fast on IoT devices, we use extracted features $\mathbf{z_x}$ instead of the raw input data $\mathbf{x}$, to represent the contextual information of input data (i.e., a state vector). 
Specifically, for the univariate data, we define the contextual state as an extracted feature vector that includes min, max, mean, and standard deviation of each day's sensor data. 
For the multivariate data, we use the encoded states of the LSTM-encoder to represent the input for the policy network. 

The policy network is trained to find an optimal policy $\pi_{\theta^*}$ that maps a state (of input) to an action (i.e., a model or layer) to maximize the expected reward of the selected action.
We train the policy network using the policy gradient \cite{williams1992REINFORCE,sutton2000policy} method to minimize the negative expected reward:
\begin{align}
\label{eq:expectedReward}
\min \mathcal{L}(\theta) = -\mathop{\mathbb{E}}_{\mathbf{a} \sim \pi_{\theta}}[R(\mathbf{a},\mathbf{z_x})],
\end{align}
where $R(\mathbf{a},\mathbf{z_z})$ is a reward function of action $\mathbf{a}$ given state $\mathbf{z_x}$.
The gradient of $\mathcal{L}(\theta)$ is derived as follows:
\begin{align}
\nabla_{\theta} \mathcal{L}
& = - \int R(\mathbf{a},\mathbf{z_x})  \nabla_{\theta} \pi_{\theta}(\mathbf{a}|\mathbf{z_x})  d \mathbf{a}  \notag  \\
& = - \int R(\mathbf{a},\mathbf{z_x})  \frac{ \nabla_{\theta} \pi_{\theta}(\mathbf{a}|\mathbf{z_x})}{\pi_{\theta}(\mathbf{a}|\mathbf{z_x})} \pi_{\theta}(\mathbf{a}|\mathbf{z_x})  d\mathbf{a}  \notag  \\
& = - \int R(\mathbf{a},\mathbf{z_x})  \nabla_{\theta} \log{\pi_{\theta}(\mathbf{a}|\mathbf{z_x})} \pi_{\theta}(\mathbf{a}|\mathbf{z_x})  d\mathbf{a}  \notag  \\
& =  -\mathop{\mathbb{E}}_{\mathbf{a} \sim \pi_{\theta}} \left[ R(\mathbf{a},\mathbf{z_x}) \nabla_{\theta} \log(\pi_{\theta}(\mathbf{a}|\mathbf{z_x}) ) \right].  \notag
\end{align}

To reduce the variance of reward value and increase the convergence rate, we utilize a reinforcement comparison with a baseline $R(\tilde{\mathbf{a}}, \mathbf{z_x})$ that is independent of output actions \cite{williams1992REINFORCE}.
We use the baseline $R(\tilde{\mathbf{a}}, \mathbf{z_x})$ as the best observed reward \cite{sutton2000policy}, which is empirically shown to boost convergence rate.
In addition, we add a $\ell_2$-norm regularization term to the loss function to prevent over-fitting problem.
So $\mathcal{L}(\theta)$ is re-written as follows:
\begin{align}
\mathcal{L}(\theta) =  -\mathop{\mathbb{E}}_{\mathbf{a} \sim \pi_{\theta}}[R(\mathbf{a},\mathbf{z_x})-R(\tilde{\mathbf{a}}, \mathbf{z_x})] + \frac{\gamma}{2} ||\theta ||_2,
\label{eq:policyGradientRegularization}
\end{align}
where $\gamma$ is a regularized parameter.
By choosing a baseline $R(\tilde{\mathbf{a}}, \mathbf{z_x})$ that is independent of output actions \cite{williams1992REINFORCE}, this can be re-written as: 
\begin{align}
\mathcal{L}(\theta)
= -\mathop{\mathbb{E}}_{\mathbf{a} \sim \pi_{\theta}}[R(\mathbf{a},\mathbf{z_x})] + R(\tilde{\mathbf{a}}, \mathbf{z_x}) + \frac{\gamma}{2} ||\theta ||_2. \notag
\end{align}

Similar to the original objective function, we minimize \eqref{eq:policyGradientRegularization} by utilizing the policy gradient method with REINFORCE algorithm \cite{williams1992REINFORCE} to compute the gradient of $\mathcal{L}(\theta)$ as follows:
\begin{align}
\nabla_{\theta} \mathcal{L} & =  -\mathop{\mathbb{E}}_{\mathbf{a} \sim \pi_{\theta}} \left[ (R(\mathbf{a},\mathbf{z_x})-R(\tilde{\mathbf{a}}, \mathbf{z_x})) \nabla_{\theta} \log(\pi_{\theta}(\mathbf{a}|\mathbf{z_x}) ) \right] + \gamma \theta \label{eq:GradientReinforceBaseline_line1} \\
& = - \mathop{\mathbb{E}}_{\mathbf{a} \sim \pi_{\theta}} \left[ \left(R(\mathbf{a},\mathbf{z_x})-R(\tilde{\mathbf{a}}, \mathbf{z_x}) \right) \nabla_{\theta} \sum_{k=1}^{K} a_k \log(s_k) \right] + \gamma \theta,
\label{eq:GradientReinforceBaseline}
\end{align}
where we substitute \eqref{eq:CategoricalDistribution} into \eqref{eq:GradientReinforceBaseline_line1} to get \eqref{eq:GradientReinforceBaseline}.
We also note that \eqref{eq:GradientReinforceBaseline_line1} is an unbiased estimator because the separated gradient term of baseline reward is zero:
\begin{align}
&  \mathop{\mathbb{E}}_{\mathbf{a} \sim \pi_{\theta}} \left[ R(\tilde{\mathbf{a}}, \mathbf{z_x}) \nabla_{\theta} \log(\pi_{\theta}(\mathbf{a}|\mathbf{z_x}) ) \right] \notag \\
= &   R(\tilde{\mathbf{a}}, \mathbf{z_x}) \mathop{\mathbb{E}}_{\mathbf{a} \sim \pi_{\theta}}   \left[  \nabla_{\theta} \log(\pi_{\theta}(\mathbf{a}|\mathbf{z_x}) ) \right]  \notag \\
= &   R(\tilde{\mathbf{a}}, \mathbf{z_x}) \int  \nabla_{\theta} \log(\pi_{\theta}(\mathbf{a}|\mathbf{z_x}) )  \pi_{\theta}(\mathbf{a}|\mathbf{z_x}) d\mathbf{a}  \notag \\
= &   R(\tilde{\mathbf{a}}, \mathbf{z_x}) \int \frac{ \nabla_{\theta} \pi_{\theta}(\mathbf{a}|\mathbf{z_x}) }{\pi_{\theta}(\mathbf{a}|\mathbf{z_x})}  \pi_{\theta}(\mathbf{a}|\mathbf{z_x}) d\mathbf{a}  \notag \\
= &   R(\tilde{\mathbf{a}}, \mathbf{z_x}) \nabla_{\theta} \int   \pi_{\theta}(\mathbf{a}|\mathbf{z_x})  d\mathbf{a}  \notag \\
= &   R(\tilde{\mathbf{a}}, \mathbf{z_x}) \nabla_{\theta} 1 = 0. \notag
\end{align}

The original REINFORCE algorithm updates policy gradient over one sample for each training step, which also causes a high variance problem.
To reduce this problem, at each training step, we use a mini-batch training of $N$ contextual states, and update the gradient by averaging over these $N$ contextual states $\mathbf{z_x}$.
The equations \eqref{eq:policyGradientRegularization} and \eqref{eq:GradientReinforceBaseline} may therefore be re-written as:
\begin{align}
\mathcal{L}(\theta) =&  -\frac{1}{N} \sum_{i=1}^{N} \mathop{\mathbb{E}}_{\mathbf{a} \sim \pi_{\theta}}[R(\mathbf{a},\mathbf{z_x}_i)-R(\tilde{\mathbf{a}}, \mathbf{z_x}_i)] + \frac{\gamma}{2} ||\theta ||_2,
\label{eq:policyGradientRegularizationBatch}\\
\nabla_{\theta} \mathcal{L} 
=& -\frac{1}{N} \sum_{i=1}^{N} \mathop{\mathbb{E}}_{\mathbf{a} \sim \pi_{\theta}} [ ( R(\mathbf{a},\mathbf{z_x}_i)
-  R(\tilde{\mathbf{a}}, \mathbf{z_x}_i) ) \nabla_{\theta} \sum_{k=1}^{K} a_k \log(s_k) ] + \gamma \theta.
\label{eq:gradientBatch}
\end{align}

In order to encourage the selection of an appropriate AD model for jointly increasing accuracy and reducing the cost of offloading tasks by pushing the chosen AD model to the edge of networks, we propose a reward function as follows:
\begin{align}
\label{eq:rewardFunction}
R(\mathbf{a}, \mathbf{z_x}) = \text{accuracy}(\mathbf{x}) -  C(\mathbf{a}, \mathbf{x}),
\end{align}
where accuracy($\mathbf{x}$) is the accuracy of detecting an input $\mathbf{x}$,
and $C(\mathbf{a},\mathbf{x})$ is the cost function of offloading the detection task to a layer $k=|\mathbf{a}|$ for data $\mathbf{x}$.
We define the cost function $C(\mathbf{a}, \mathbf{x})$ as a function that maps the end-to-end detection delay $t_\text{e2e}(\mathbf{x}, \mathbf{a})$ to an equivalent accuracy in the range $[0,1]$ with intuition that a higher delay will result in a greater reduction of accuracy:
\begin{align}
\label{eq:costFunction}
C(\mathbf{a}, \mathbf{x}) = \frac{\alpha \cdot t_\text{e2e}(\mathbf{x}, \mathbf{a}) }{1+ \alpha \cdot t_\text{e2e}(\mathbf{x}, \mathbf{a})},
\end{align}
where $\alpha$ is a tuning parameter used to tradeoff between the end-to-end delay and the accuracy (through the reward function \eqref{eq:rewardFunction}).
For example, $C(\text{Edge}, \mathbf{x})|_{t_\text{e2e}=250\,\text{ms}}=0.1$ means the end-to-end detection delay cost of offloading a sample $\mathbf{x}$ to an edge server 250\,ms is equivalent to a reduction 0.1 in accuracy.
The end-to-end delay $t_\text{e2e}(\mathbf{x}, \mathbf{a})$ consists of the communication delay
of transmission data $\mathbf{x}$ from an IoT device to a server at the layer $k=|\mathbf{a}|$ of HEC,
and the computing delay
of executing the detection task at the layer $k$.

We summarize the procedure of training policy network as Algorithm \ref{alg:REINFORCE}.

To balance between exploration and exploitation during training, we apply a decayed-$\epsilon$-greedy algorithm for action selection.
We train the policy network over a number of epochs $N_\text{epochs}$ with an initial exploration probability of $p_i$ that decays over $A$ steps to a final value of $p_e$.
So in each training episode, the actual $\epsilon$ is calculated as Lines~\ref{alg:epsilon} and \ref{alg:epsilonEnd} of Algorithm \ref{alg:REINFORCE}.
Then, an action is randomly selected to explore with probability $\epsilon$, while with probability $(1-\epsilon)$ an action is greedily selected based on output of the current policy network.

\begin{algorithm}[!t]
    \SetKwInput{KwInput}{Input}
    \SetKwInput{KwOutput}{Output}
    \SetAlgoLined
    \KwInput{
        contextual states $\mathbf{z_{x}} \in \mathbf{Z_{x}}$, policy network $\pi_{\theta}$,\\
         learning rate $\alpha$, regularizer $\gamma$,
        $N_\text{epochs}$,\\
        $A$, \footnotesize{\tcp{Number steps of decayed-$\epsilon$ for expoloration}}
        $p_i \in [0,1]$, \footnotesize{\tcp{Initial exploration probability}}
        $p_e \in [0,1]$. \footnotesize{\tcp{Ending exploration probability}}
    }
    \KwOutput{trained policy $\pi_{\theta^*}$. }
    Initialize policy $\pi_{\theta}$, $n_e \gets 0$\;
    \While{$n_e < N_\text{epochs}$}{
        Select a mini-batch of size N of $\mathbf{z_{x}}$\;
        Calculate the current output of policy $\pi_{\theta}$\;
        $r = \max\left(\frac{A - n_e}{A}, 0\right)$\; \label{alg:epsilon}
        $\epsilon \gets (p_i - p_e) r + p_e$\; \label{alg:epsilonEnd}
        Select actions $\mathbf{a}$ based on decayed-$\epsilon$-greedy method\;
        \label{alg:epsilonGreedy}
        Getting accuracies and end-to-end delays from corresponding AD models\;
        \label{alg:GettingAccDelayFromEvironment}
        Calculate costs as \eqref{eq:costFunction} and rewards $R(\mathbf{a},\mathbf{z_x})$\;
        \label{alg:CalculateCost}
        Calculate gradient according to \eqref{eq:gradientBatch}\;
        Update the parameters of network $\theta \gets \theta - \alpha \nabla_{\theta} \mathcal{L}(\theta) $\;
        Update baseline $R(\tilde{\mathbf{a}},\mathbf{z_x})$\;
        \label{alg:update_paramters}
        $n_e \gets n_e+1$\;
    }
    \label{alg:end_while_train}
    \caption{REINFORCE algorithm}
    \label{alg:REINFORCE}
\end{algorithm}

\section{Implementation \& Experiment Setup}
\label{sec:ExpSetup}
\subsection{Dataset}
We evaluate our proposed approach with two public datasets.
The data is standardized to zero mean and unit variance for all of the training tasks and datasets.

\begin{figure}[htb]
    \centering
    \subfloat[Normal week]{
        \includegraphics[width=0.5\linewidth]{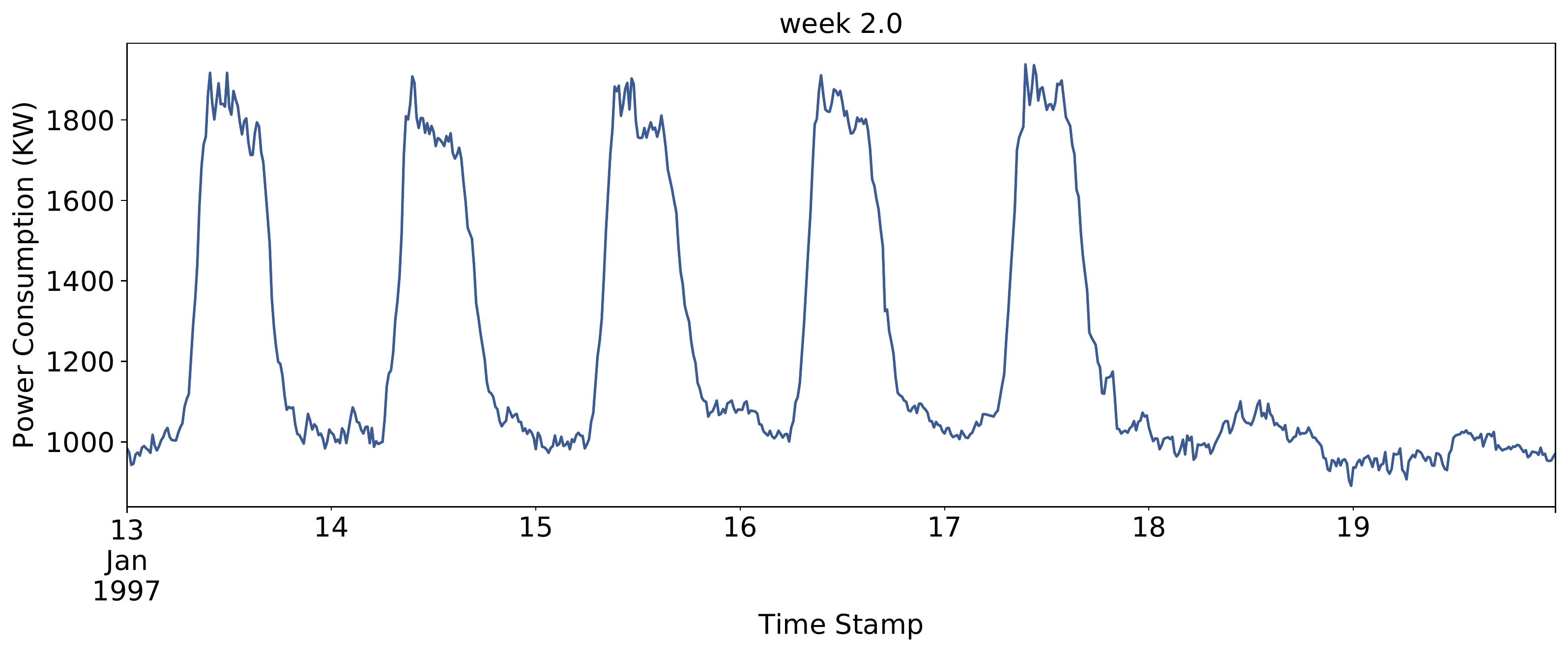}%
        \label{fig:PowerDemandNormalWeek}%
    } 
    \subfloat[Abnormal week]{
        \includegraphics[width=0.5\linewidth]{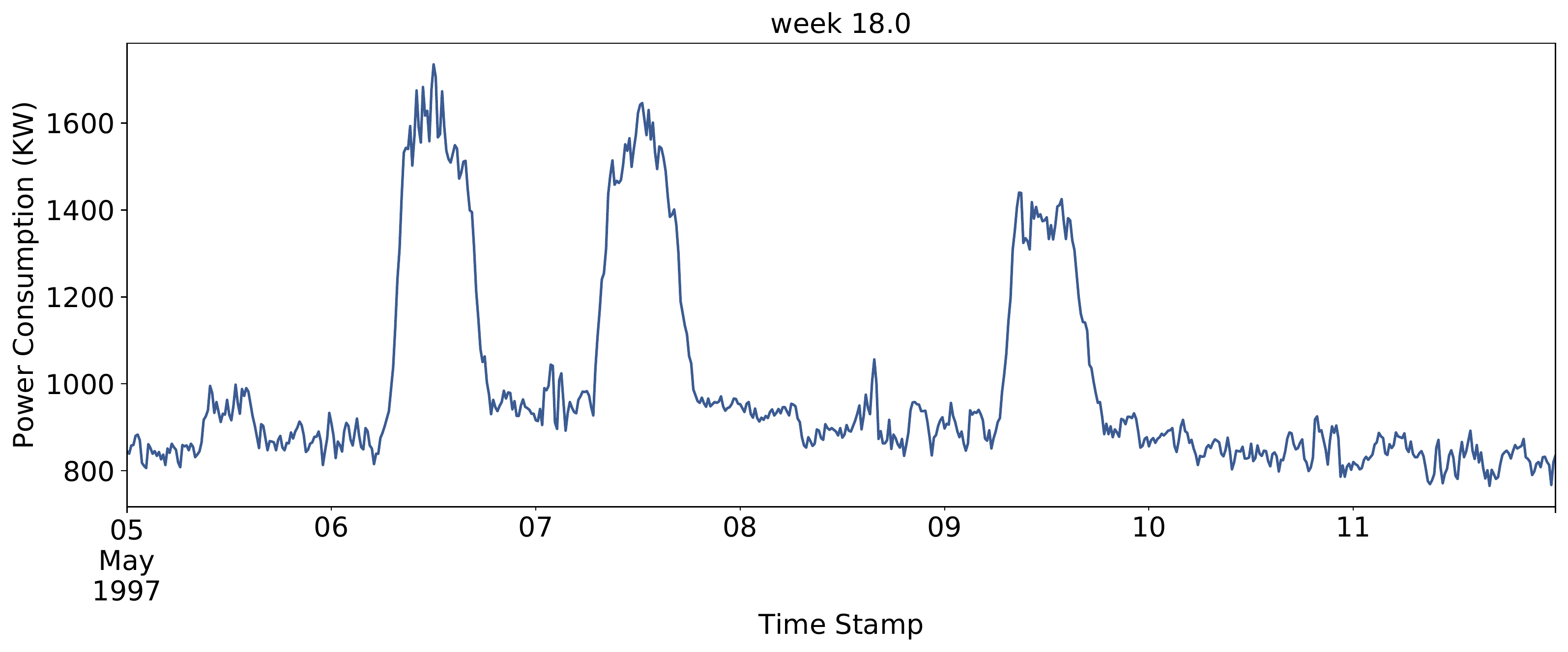}%
        \label{fig:PowerDemandAbnormalWeek}%
    }
    \caption{Example weeks of univariate dataset (i.e., power consumption).}
    \label{fig:ExamplePowerDemand}
\end{figure}

\textbf{Univariate dataset.} We use a dataset of power consumption\footnote{\url{http://www.cs.ucr.edu/~eamonn/discords/}} of a Dutch research facility 
that has been used in \cite{Keogh_HotSax_ICDM2005,Malhotra_LSTM_encDec_ICMLWrsh2016,singh2017anomaly}.
It {comprises 52 weeks of power consumption and} consists of 35040 samples recorded every 15 minutes.
The data has a repetition of weekly cycle of 672 time steps with five consecutive peaks for five weekdays and two lows for weekends.
Abnormal weeks could have less than five consecutive peak days on weekdays (perhaps due to a holiday), or high power consumption on a weekend.
Examples of normal and abnormal weeks are shown in Fig.~\ref{fig:ExamplePowerDemand}.
Hence, each input data is a sequence of one week of data with 672 time steps.
We manually label a day as abnormal if it is a weekday with low power consumption, or it is a weekend with high power consumption; other days are labeled as normal.
For the AD task, we split the dataset into train and test sets with ratio 70:30, or equivalently 37 weeks:15 weeks.
The training set only contains normal weeks and the test set contains the remaining normal weeks and all the 8 anomalous weeks, each having at least one abnormal day.
For training the policy network, we choose a training set that contains all the 8 abnormal weeks and 7 normal weeks, and a test set that is the whole dataset to verify the quality of the policy network.

\begin{figure}[htb]
    \centering
    \subfloat[Normal window (Walking).]{
        \includegraphics[width=0.5\linewidth]{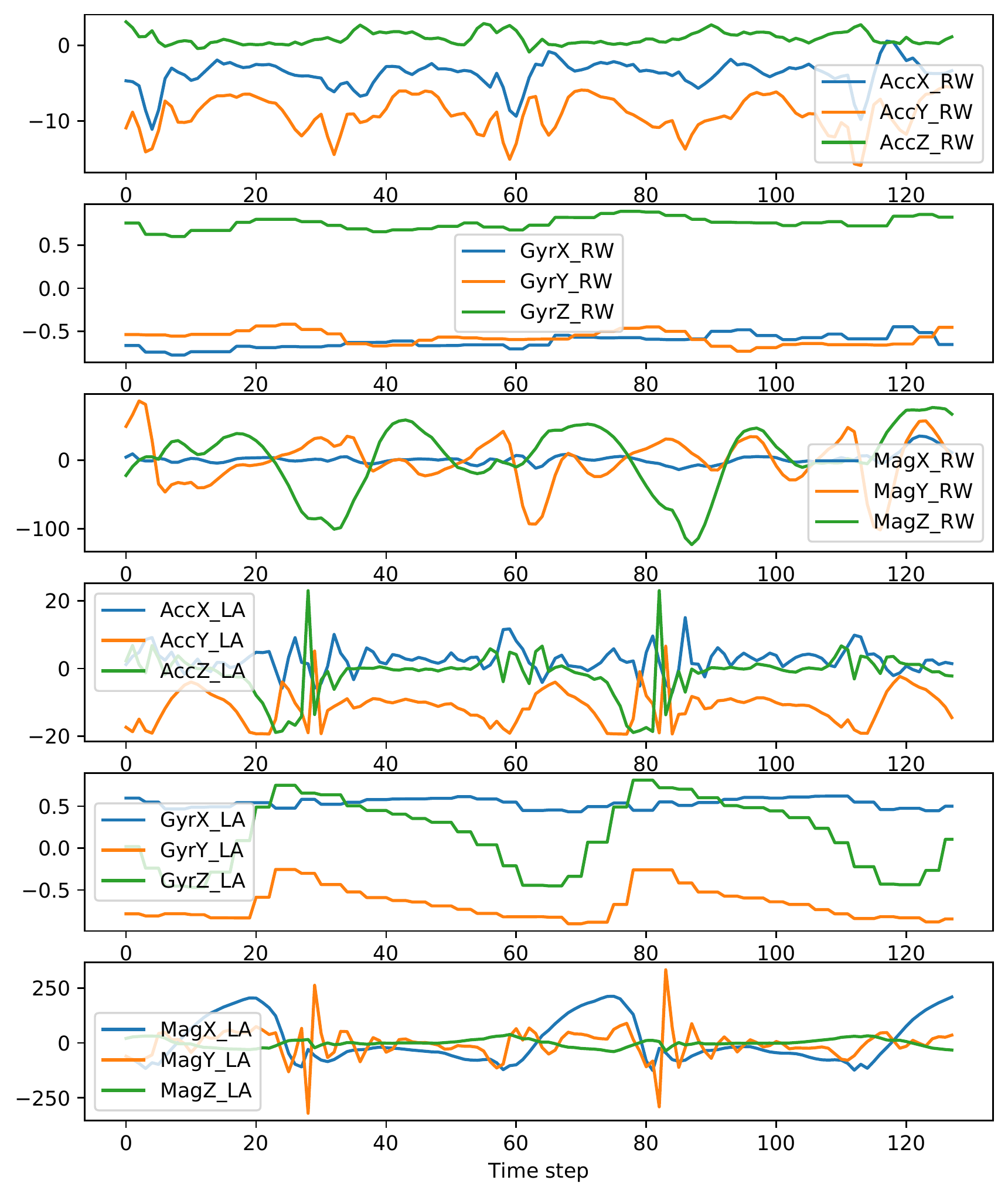}%
        \label{fig:MhealthNormalWindow}%
    }
    \subfloat[Abnormal window (Cycling).]{
        \includegraphics[width=0.5\linewidth]{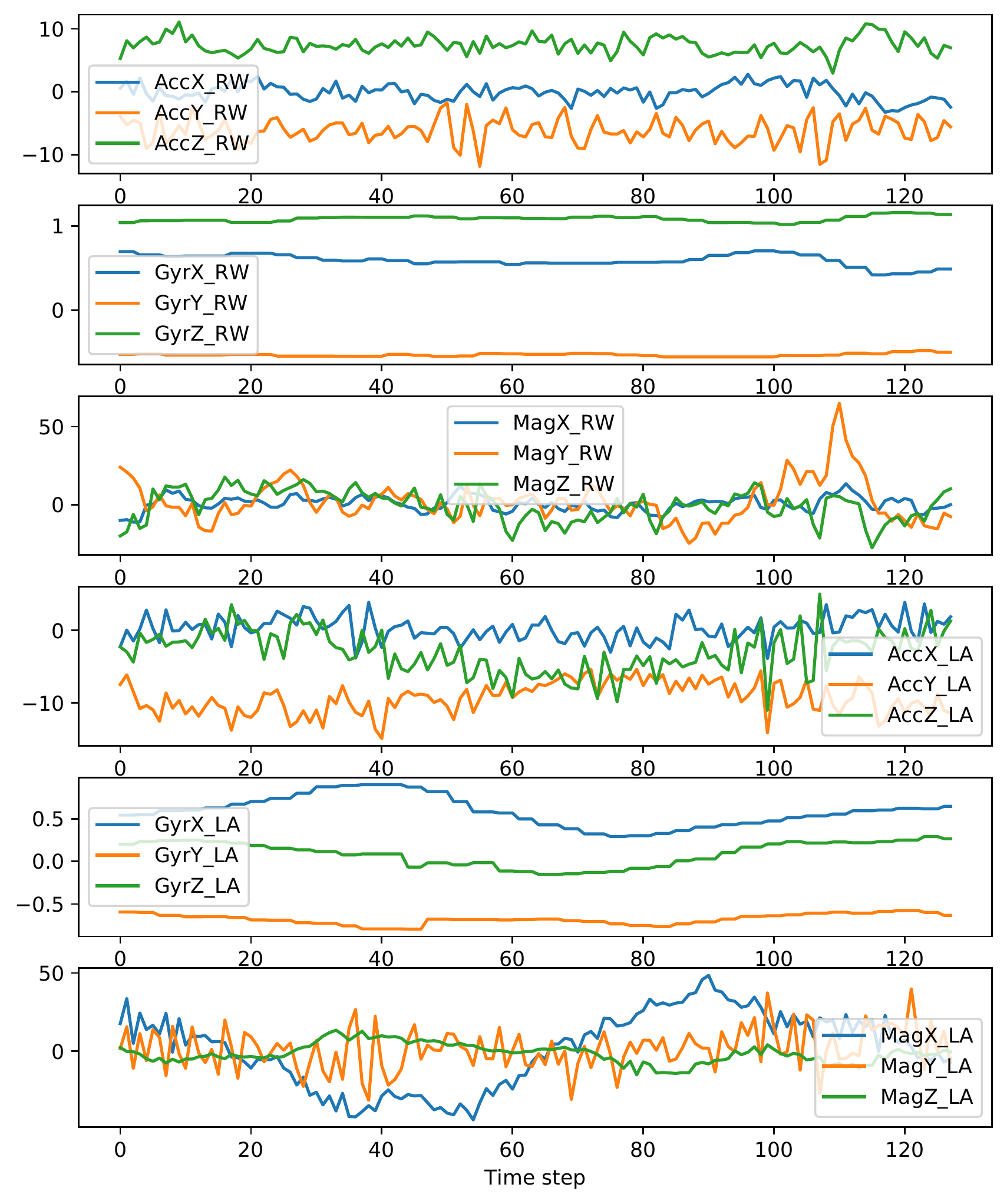}%
        \label{fig:MhealthAbnormalWindow}%
    }
    \caption{Example time windows of multivariate dataset (i.e., MHEALTH).}
    \label{fig:ExampleMhealth}
\end{figure}

\textbf{Multivariate dataset.}
We use MHEALTH dataset\footnote{\url{http://archive.ics.uci.edu/ml/datasets/mhealth+dataset}} which consists of 12 human activities of 10 different people.
And each person wore two motion sensors: one on left-ankle and the other on right-wrist.
Each motion sensor contains a 3-axis accelerator, a 3-axis gyroscope, and a 3-axis magnetometer; hence the input data has 18 dimensions.
The sampling rate of these sensors is 50\,Hz.
We use a window sequence of 128 time-steps ($\sim$2.56 second) with a step-size of 64 between two windows.
Adopting the common practice, we choose the dominant human activity (e.g., walking) as normal and treat the other activities as anomalous.
Fig.~\ref{fig:ExampleMhealth} shows an example of normal and abnormal window sequences.
For the AD task, we select 70\% of normal samples of all the subjects (people) as the training set and the remaining 30\% of normal samples plus 5\% of each of the other activities as the test set.
To train the policy network, we select 30\% of normal samples and 5\% of each of the other activities as the training set, and the whole dataset as the test set.

\subsection{Implementation of Anomaly Detection Models and Policy Network}
\label{subsec:Implementation}

We use Tensorflow and Keras to implement the AD models\footnote{We can package each AD model in a Docker container, a light-weight virtual computing software environment, and run the container within an edge server or cloud server, which will make deployment of edge computing systems much easier and more scalable, as studied in \cite{La_FogComputing_2019, Ngo_Globecom2020}.} (i.e., three AE models and three LSTM-seq2seq models as shown in Fig.~\ref{fig:OverviewSystem}) and the policy network model.
For the univariate time-series dataset, the input is a sequence of 672 time steps---a week of measured power consumption.
We use \texttt{tanh} as the non-linear activation function for all the hidden layers, and the linear function for output layer of the autoencoder models.
To prevent the overfitting problem, we apply dropout technique with rate 30\% during training, and add a $\ell_2$-regularization term.
We train and test the three models separately with $5$-fold cross-validation.

For the multivariate time-series dataset, the input is a sequence of 128 time steps of 18-dimensional data.
The LSTM-seq2seq-IoT model, which will be deployed on Raspberry Pi 3, consists of 50 vanilla LSTM units for each encoder and decoder.
Since the edge server and cloud server are empowered with GPU, we implement the LSTM-seq2seq-Edge and BiLSTM-seq2seq-Cloud models based on \texttt{CuDNNLSTM} units to accelerate training and inference time.
Before deploying the LSTM-seq2seq-IoT and LSTM-seq2seq-Edge models on Raspberry Pi3 and Jetson-TX2, we compress them by (i) removing the trainable nodes from the graph and converting variables into constants; (ii) quantizing the model parameters from floating-point 32-bit (FP32) to FP16.
We observe no performance decrease of these compressed AD models, while the average inference delay of these compressed AD models running on Raspberry Pi3 and Jetson-TX2 are reduced by 61\,ms and 126.1\,ms, respectively.

The policy network requires low complexity and needs to run fast on IoT devices without consuming enough resources to affect to the IoT detection task.
So the state input to the policy network needs to be small but still well represent the whole sequence of input data.
For the univariate data, we define the contextual state as an extracted feature vector which includes min, max, mean, and standard deviation of each day's sensor data.
Thus the dimension of the contextual state is just 4x7=28.
For the multivariate data, we use the encoded states of the LSTM-encoder to represent the input state for the policy network; hence, the dimension of the contextual state, which is the encoded states constructed by concatenation of (h,c), is 50 + 50 =100.
Subsequently, we build the policy network as a single hidden neural network (with 100 and 300 hidden units for univariate and multivariate data, respectively) and a softmax layer with 3 output units, i.e., $\sigma(o_i)=\frac{\exp(o_i)}{\sum_{j=1}^{K}\exp(o_j)}$, which indicate the likelihood of choosing one of three AD models.
We train the policy network as described in Section \ref{subsec:DynamicModelSelectionScheme} with 6000 and 600 episodes\footnote{The univariate dataset needed more episodes than the multivariate dataset because the size of univariate dataset is smaller (even though we repetitively replay the training set with randomly shuffling the input sequences) and, as a result, the convergence of the reward value of the policy network was observed to be slower. In addition, the convergence rate is also dependent on the parameter $\alpha$, and we set the number of training episodes to the maximum corresponding to the worst $\alpha$.} for univariate and multivariate datasets, respectively; and the initial $\epsilon=0.5$ is gradually decreased to zero after half the number of episodes.
We empirically select $\alpha=0.0005$ (as shown in Section~\ref{subsec:Alpha}) for both univariate and multivariate datasets to calculate the cost of executing detection as given by \eqref{eq:costFunction}.

\subsection{Accelerated Training of Policy Network}
\label{subsec:AcclerationTrainingPolicyNetwork}
Recently, Google has published a distributed mechanism to efficiently accelerate the training process for deep reinforcement learning problems \cite{Espeholt2019Seed}.
Inspired by this work, we can accelerate our training policy network with mini-batch training as in Algorithm~\ref{alg:REINFORCE} by modifying Line~\ref{alg:GettingAccDelayFromEvironment} as follows:
(i) instead of sequentially querying reward of each input sample, we group inputs that belong to the same action and send them to the appropriate AD model to do parallel inference in a batch manner;
(ii) 
we also can concurrently do inference from multiple AD models at $K$ layers of HEC if there are more than one AD model in the action outputs of the $\epsilon$-greedy method. 

With this proposal (called the \textit{parallel} approach), we expect to reduce time to train policy network because of
(1) reduction of communication overhead between the training server and multiple AD models at multiple layers of HEC, and
(2) leveraging mini-batch inference at each AD model instead of one single input detection.
The results will be analyzed in Section \ref{subsec:ResultAcceleration}.
Note that, when training the parallel approach, the detection delay is measured for multiple samples due to concurrently querying multiple samples.
Therefore, we use the average of the end-to-end detection delays for each HEC layer that were collected from the normal training approach to calculate rewards. 

\subsection{Software Architecture and Experiment Setup}

\begin{figure}[tb]
    \centering
    \includegraphics[width=0.8\linewidth]{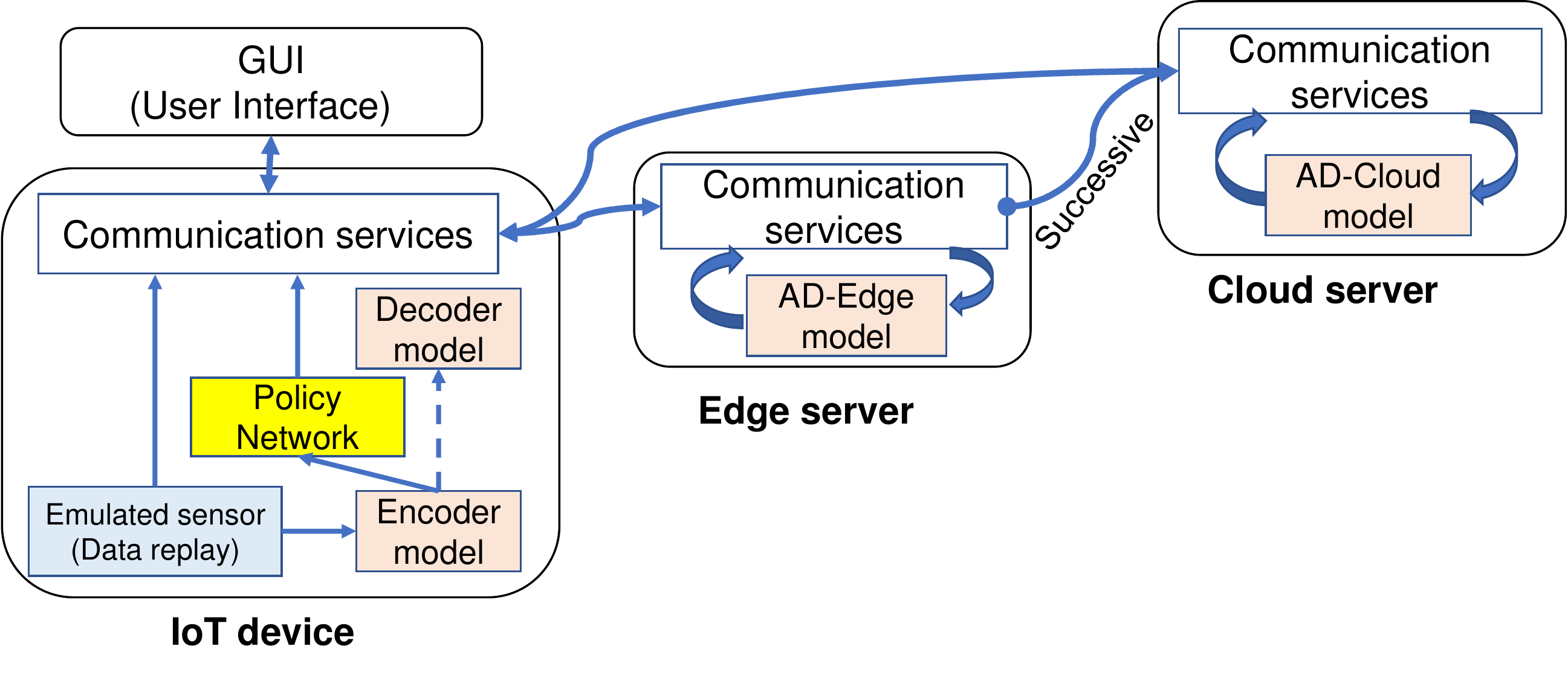}%
    \caption{Software architecture of our HEC system which consists of different AD models at IoT device, edge server and cloud server, and a contextual-bandit policy network residing at the IoT device.}
    \label{fig:SoftwareArchitecture}
\end{figure}

\begin{figure}
    \centering
    \subfloat[GUI: select dataset and evaluation scheme, and adjust parameters.]{
        \includegraphics[width=0.7\linewidth]{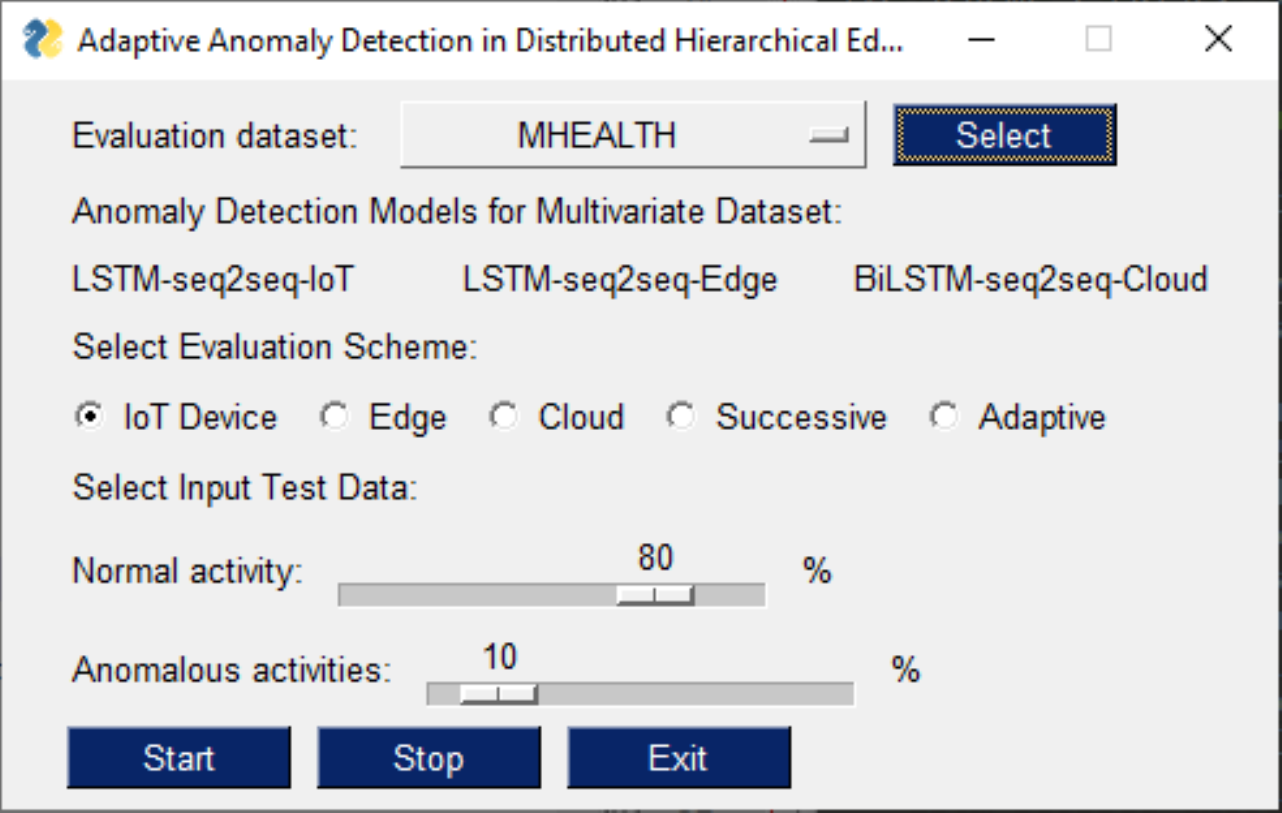}%
        \label{fig:DemoInterface}
    }

    \subfloat[Result panel: raw sensory signals, AD performance and associated actions.]{
        \includegraphics[width=0.85\linewidth]{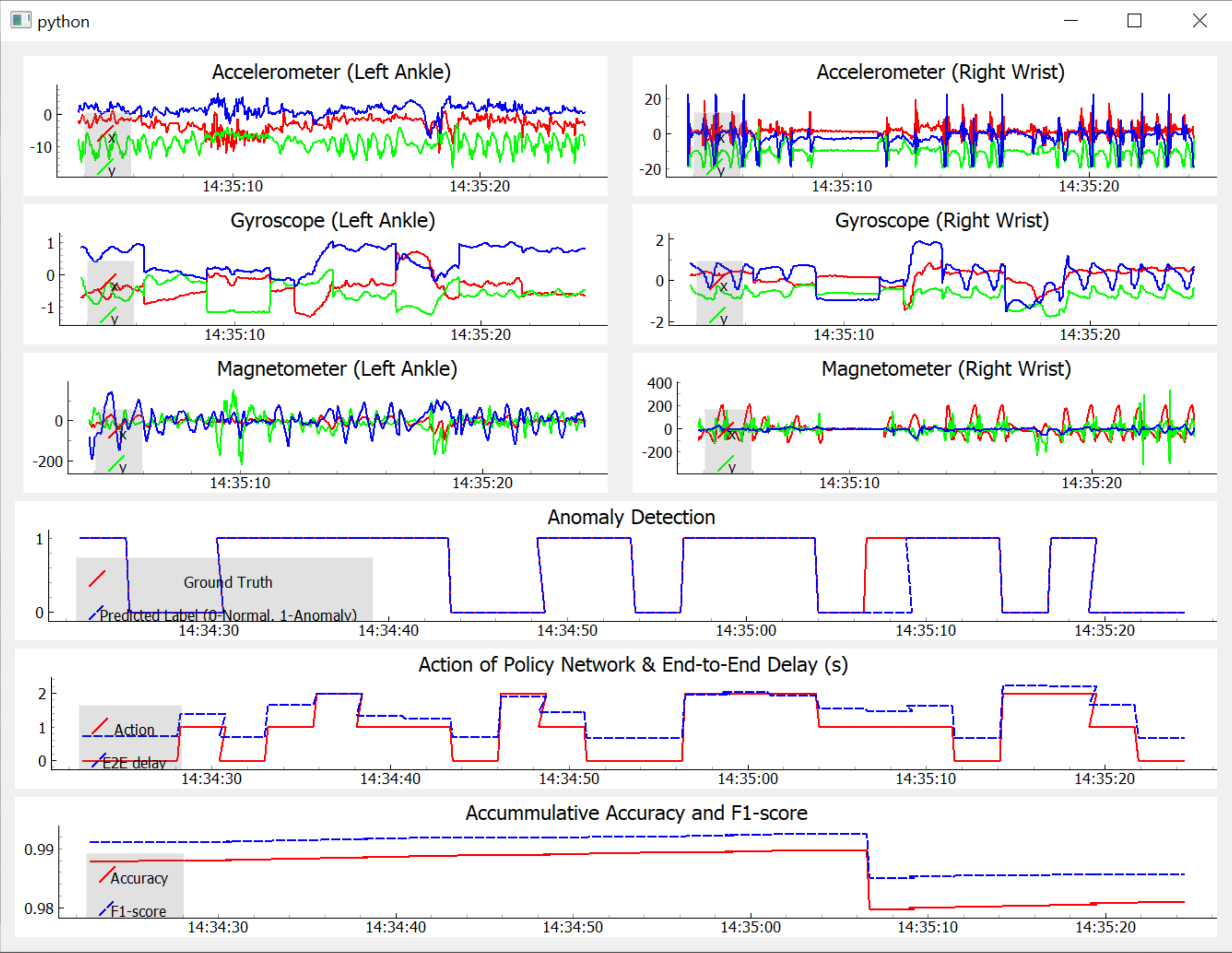}%
        \label{fig:DemoGraph}
    }
    \caption{GUI and multivariate results.}
    \label{fig:DemoGUI}
\end{figure}

\begin{figure}[t]
    \centering
    \includegraphics[width=1.0\linewidth]{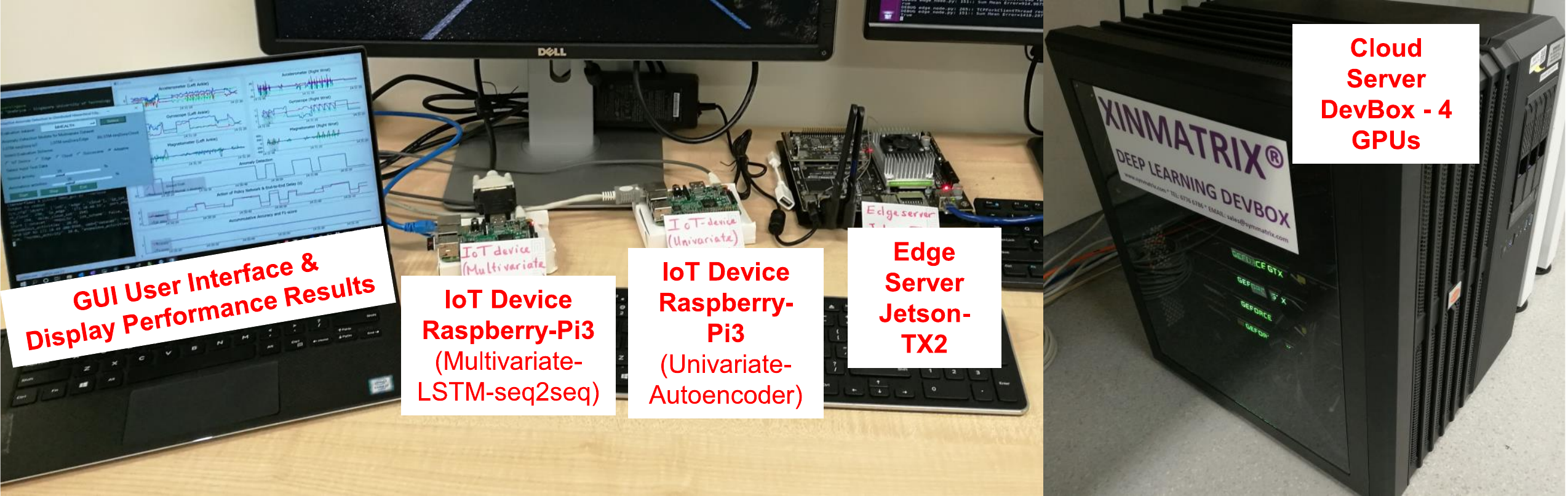}%
    \caption{Hardware setup of our HEC testbed.}
    \label{fig:HardwareTestbed}
\end{figure}

The software architecture of our HEC system\footnote{A brief introduction of our HEC testbed is also available online: \url{https://rebrand.ly/91a71}} is shown in Fig.~\ref{fig:SoftwareArchitecture}.
It consists of a GUI, the adaptive model selection scheme based on the policy network, and the three AD models at the three layers of HEC.
The GUI allows a user to select which dataset and model selection scheme to use, as well as to tune parameters, as shown in Fig.~\ref{fig:DemoInterface}, and displays the sensory raw signals and performance results, as shown in Fig.~\ref{fig:DemoGraph}.
All the communication services use keep-alive TCP sockets after opening in order
to reduce the overhead of connection establishment.
Network latency (round-trip time) as shown in Fig.~\ref{fig:OverviewSystem} is configured by using the Linux traffic control tool, \texttt{tc}, to emulate the WAN connections in HEC.
The hardware setup for the HEC testbed is shown in Fig.~\ref{fig:HardwareTestbed}.
To emulate an environment with high-speed incoming IoT data that requires fast anomaly detection, we replay the datasets with increased sampling rates.
For the univariate dataset (power consumption), we replay it with a sampling rate of 672 samples per second during the experiments, as compared to the dataset's original sampling rate which is one sample per 15 minutes.
As such, a whole year of power consumption data was replayed and processed within minutes in our experiments.
For the multivariate dataset (healthcare), we replay the sensory data with a simulated sampling rate of 128 samples per second, as compared to the original sampling rate 50 samples per second.

We measure end-to-end delays $t_\text{e2e}(\mathbf{x}, \mathbf{a})$ on actual IoT devices, which is the interval between the starting time when a sample input sequence is generated at an IoT device, and the end time when the detection result is received at the IoT device.
Note that the actual anomaly detection was executed on exactly one of the three layers (IoT, edge, or cloud).
Based on the measured $t_\text{e2e}$, we calculate the cost of detection using \eqref{eq:costFunction}.
We will evaluate with different parameters $\alpha$ to see the trade-off between the offloading cost and the accuracy gain of a complex model.

\textbf{User Actions:}
As shown in Fig.~\ref{fig:DemoInterface}, we allow users to interact and evaluate the HEC testbed performance with (i) either univariate or multivariate datasets, (ii) different fractions of normal and abnormal data in the datasets to use, and (iii) different schemes under evaluation:
(1) the \textit{IoT Device scheme}, which always detects directly on IoT devices,
(2) the \textit{Edge scheme}, which always offloads to an edge server,
(3) the \textit{Cloud scheme}, which always offloads to the cloud,
(4) the \textit{Successive scheme}, which first executes on IoT devices and then successively offloads to a higher layer until reaching a confident output or the cloud, or finally
(5) the \textit{Adaptive scheme}, which is our proposed adaptive model selection scheme. 
After the user clicks ``Start'', our result panel as in Fig.~\ref{fig:DemoGraph} will show the continuously updated raw sensory data (accelerometer, gyroscope, magnetometer), anomaly detection outcome (0 or 1) vs. ground truth, detection delay vs. the actions determined by our policy network, and the accumulative accuracy and F1-score.

\subsection{State-of-the-Art Schemes in Comparison}
\label{subsec:STOAschemes}
Besides comparing our Proposed scheme with the baseline schemes (i.e., IoT Device, Edge, Cloud and Successive schemes) described above, we compare our scheme with the state-of-the-art methods (referred to as \textit{kNN-sequence} \cite{Taylor_SIGPLAN2018}, and \textit{Adapted-BlockDrop} \cite{Blockdrop_WuCVPR2018}).
In the kNN-sequence \cite{Taylor_SIGPLAN2018}, a series of light-weight kNN classifiers are used to provide decisions for choosing a proper inference model to use.
We can directly apply this method with our datasets and our trained AD models by implementing three kNN classifiers (with $k=4$) for univariate data, and three kNN classifiers (with $k=3$) for multivariate data.
Note that, instead of using $k=5$ as the paper \cite{Taylor_SIGPLAN2018}, we choose $k$ based on grid search for each dataset to achieve the best performance results.

Besides {\it kNN-sequence}, we also implement a variant of \cite{Taylor_SIGPLAN2018}, called \textit{kNN-single}, as another basis for comparison. It directly provides a selected layer for detecting anomaly based on extracted features of the input IoT data.
For both kNN-sequence and kNN-single schemes, we adopt a similar training procedure from \cite{Taylor_SIGPLAN2018} to generate training data (i.e., feature inputs and output classes) and build the classifier models.

The prior work \cite{Blockdrop_WuCVPR2018} (i.e., BlockDrop) is strongly tied with the computer vision ResNet architecture based on multiple stacked CNNs blocks, which can be bypassed by identity skip-connections to reduce complexity for each image-specific input.
In contrast, our paper tackles sequential time-series data and allows to use different model architectures for different complexities, resulting in more flexibility for designing AD models at HEC layers.
In terms of similarity, our approach and \cite{Blockdrop_WuCVPR2018} both use a policy network to make decisions, but with (i) different input (extracted feature of a sequential window vs. an image) and (ii) different output (a selected HEC's layer for inference vs. a binary vector for choosing which CNN blocks involving in a reduced ResNet), (iii) different reward functions, and (iv) with/without network latency in mind.
In addition, the policy network in BlockDrop (which also uses a ResNet architecture) is not light-weight and is therefore unsuitable to run on IoT devices in the HEC scenario.
To make a fair comparison with BlockDrop \cite{Blockdrop_WuCVPR2018} in the anomaly detection task, we have followed their approach by implementing a policy network using the same extracted feature of sequential input, the same type of output as our application, and the same two fully-connected layers architecture as our approach.
With this setup, we train this mimicking policy network, namely \textit{Adapted-BlockDrop}, with the reward function and parameters $\gamma=[2, 5, 10]$ as in \cite{Blockdrop_WuCVPR2018} to penalize incorrect prediction at IoT Device, Edge, and Cloud layers, respectively.

The idea of splitting large neural networks into smaller portions for each layer of HEC, such as in BranchyNet \cite{Teerapittayanon_ICDCS2017} and Neurosurgeon \cite{Neurosurgeon_Kang2017}, is an interesting approach for image classification task and can save network bandwidth by transmitting compressed intermediate results between portions.
However, these partial models must be portions of the same DNN and be jointly trained together with a combined objective function, while our approach has the flexibility to use independent models at different HEC layers, taking advantage of any available state-of-the-art DNN architecture that is best suited for each layer.
In addition, BranchyNet was designed for CNNs, and its implementation involved splitting in the forward path and carefully gathering gradient losses in the backward path during backpropagation training. Transforming its idea of splitting and stacking to RNN (such as LSTM or GRU, as in our case) is not trivial and is worth another new study.
Therefore, we do not directly compare with BranchyNet.
On the other hand, BranchyNet \cite{Teerapittayanon_ICDCS2017} is similar to the \textit{Successive} scheme (which we compare) in the following sense: BranchyNet attempts its deep CNNs portions in a successive manner, and if an early portion does not satisfy the performance requirement, it will move on to the next portion of the deep CNNs.

\section{Experiment Results}
\label{sec:ExperimentResults}
\subsection{Comparison Among Anomaly Detection Models}
\label{subsec:ComparisonADModels}
\begin{figure}[!htb]
    \centering
    \subfloat[Absolute reconstruction errors.]{
        \includegraphics[width=0.75\linewidth]{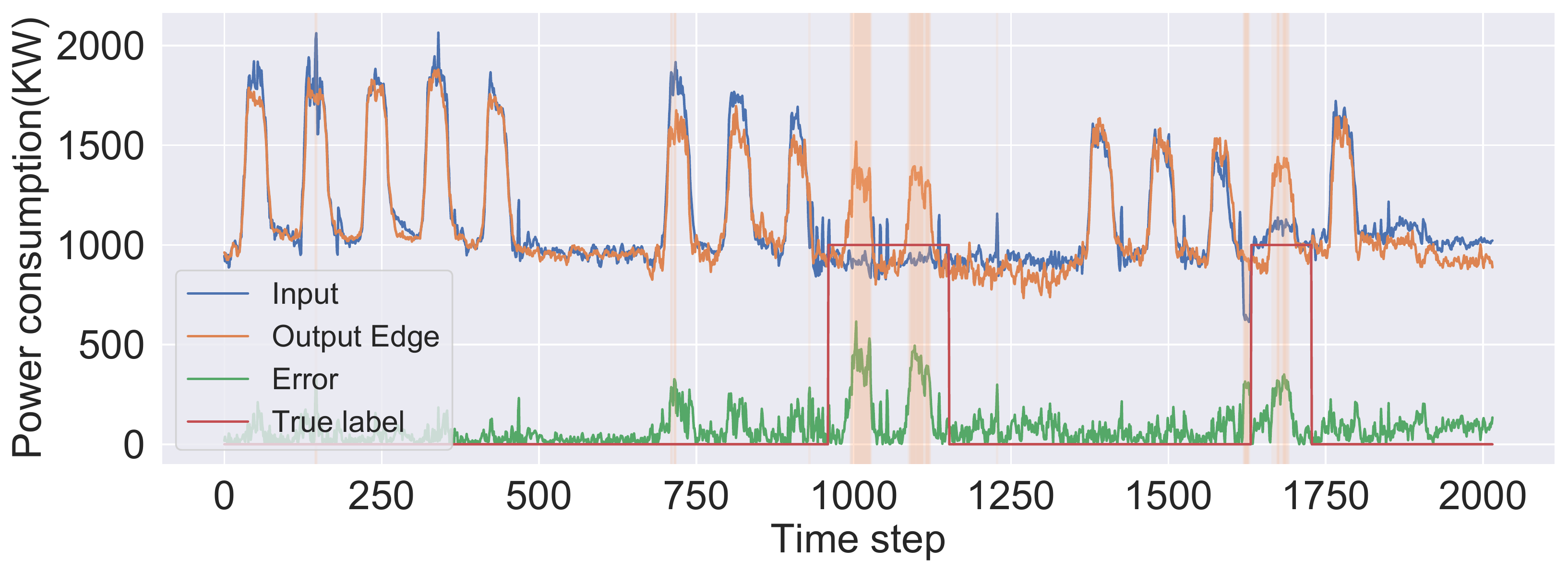}%
        \label{fig:TestReconstructionLossEdge}%
    } \quad
    \subfloat[Logarithmic probability densities (logPD) of errors.]{
        \includegraphics[width=0.75\linewidth]{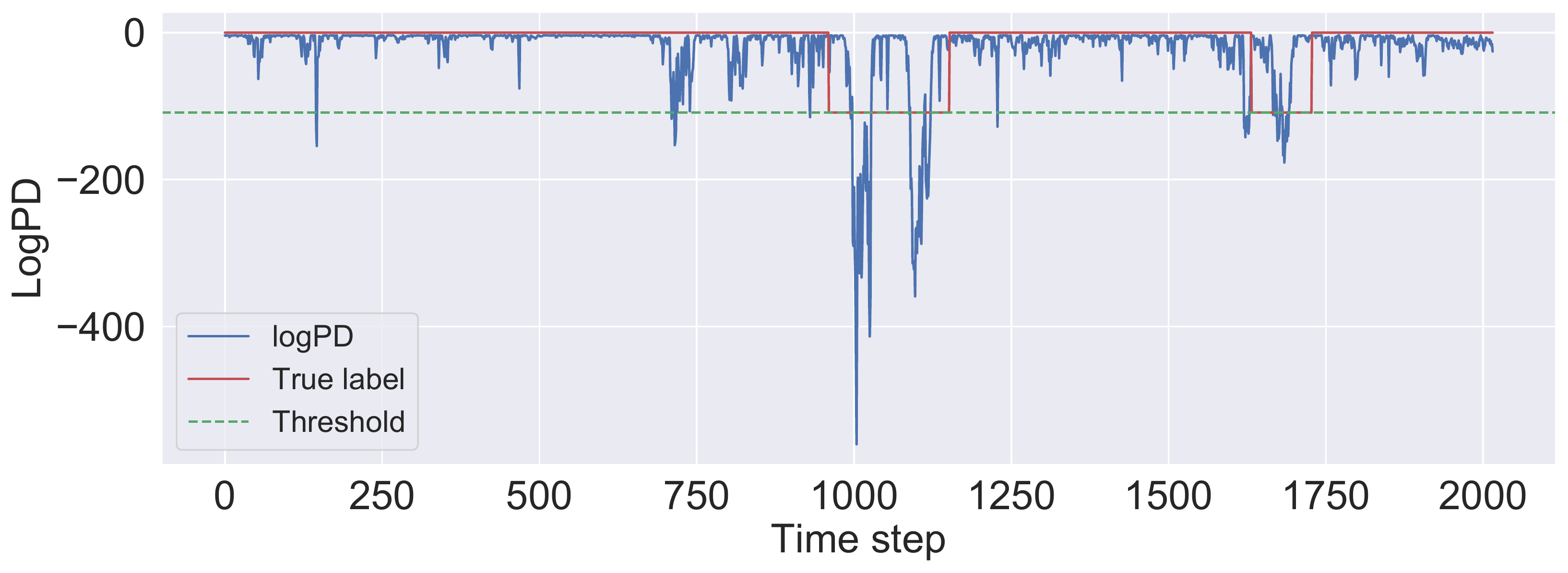}%
        \label{fig:TestLogPDEdge}%
    }
    \caption{AE-Edge performance on test set for univariate data.}
    \label{fig:ResultEdgeModel}
\end{figure}

Table~\ref{tab:threeAnomalyDetectors} compares the performance and complexity among the three AD models we use.
For both univariate and multivariate data, the complexity of AD models increases from IoT to cloud, as indicated by ``\# of Parameters'' (weights and biases) which reflects the approximate {\em memory footprint} of the models, and total ``FLOP'' (floating-point operations) which reflects the required computation of each model during the inference.
Along with this, the F1-score and accuracy increase as well; for example, these metrics are 95.5\% and 19\% higher for AE-Cloud than for AE-IoT, and 15\% and 18.9\% higher for BiLSTM-seq2seq-Cloud than for LSTM-seq2seq-IoT.
To illustrate the nature of the edge model errors, we show an example of reconstruction performance of the AE-Edge model in Fig.~\ref{fig:ResultEdgeModel}.

On the other hand, the execution time (for running the detection algorithms) decreases from IoT to cloud, as indicated in the last row of Table~\ref{tab:threeAnomalyDetectors}, which is measured on the actual machines of our HEC testbed and averaged over five runs.
This is due to the different computation capacity (whereas communication capacity is taken into account by our end-to-end delay shown later).
One more observation is that LSTM-seq2seq models, which handle multivariate datasets, take much longer time to run (up to 591 ms) than AE models, which handle univariate datasets (up to 12.4 ms).

\begin{table}[ht]
    \caption{Comparison among AD models.}
    \label{tab:threeAnomalyDetectors}
    \centering
    \begin{tabular}{ l@{\hspace{0.8em}} l@{\hspace{0.8em}} l@{\hspace{0.8em}} l@{\hspace{0.8em}}  l@{\hspace{0.8em}} l@{\hspace{0.8em}} l@{\hspace{0.8em}} }
        \toprule
        \textbf{Dataset/Model} & \multicolumn{3}{l}{\textbf{Univariate/Autoencoder}} & \multicolumn{3}{l}{\textbf{Multivariate/LSTM-seq2seq} }\\
        \midrule
        \textbf{Layer} & \textbf{IoT} & \textbf{Edge} & \textbf{Cloud} & \textbf{IoT} & \textbf{Edge} & \textbf{Cloud} \\
        \midrule
        \textbf{\#Parameters}       & 271,017 & 949,468 & 1,085,077 & 28,518 & 97,818 & 1,028,018 \\
        \textbf{FLOP}               & 1.35M   & 2.93M & 5.41M      & 3.92M  & 7.84M  & 31.33M \\ 
        \textbf{Accuracy(\%)}      & 78.09   & 93.33 & 98.09      & 82.63 & 94.21 & 97.37 \\
        \textbf{F1-score}          & 0.465   & 0.741 & 0.909     & 0.852 & 0.955 & 0.980 \\
        \textbf{Exec time (ms)}    & 12.4    & 7.4  & 4.5        & 591.0  & 417.3 & 232.3 \\ 
        \bottomrule
    \end{tabular}
\end{table}


\subsection{Comparison Among Model Selection Schemes}
\label{subsec:ComparisonModelSelectionScheme}

The F1-score, accuracy, average detection delay, and total reward of the entire univariate and multivariate datasets under four baseline schemes, three state-of-the-art schemes, and our proposed scheme are shown in Table \ref{tab:ExperimentResultDynamicScheme}.
We can see that the IoT Device scheme achieves the lowest detection delay but also the poorest accuracy and F1-score among all the evaluated schemes.
On the other extreme, the Cloud scheme yields the best accuracy and F1-score but incurs the highest detection delay (end-to-end).
The Successive scheme leverages distributed anomaly detectors in HEC and thus significantly reduces the average detection delay as compared to the Edge and Cloud schemes.
However, its accuracy and F1-score are outperformed by the Edge scheme.
In contrast, our proposed adaptive scheme adaptively selects a suitable model to execute the AD task to jointly maximize accuracy and minimize detection delay.
Thus, not only does it achieve lower detection delay but its F1-score and accuracy also consistently outperform those of IoT Device, Edge, and Successive schemes.
For univariate data, even though the F1-score and accuracy of our proposed scheme are marginally lower than those of the Cloud scheme by 4.3\% and 0.35\% respectively;
our scheme reduces the end-to-end detection delay by a substantial 84.9\%.
For multivariate data, we got an interesting result (with $\alpha=0.0005$) that the proposed scheme outperforms the Cloud scheme not only in terms of delay (reduction by 10.6\%), but also in F1-score and accuracy.
We believe that in this case our proposed scheme of the multivariate data outperforms the Cloud scheme in F1-score and accuracy because: 
(i) \textit{quality}: the contextual information of input data (i.e., encoded states from LSTM-encoder) is able to well capture feature representations of the multivariate data which help policy network deliberately choose the best destination AD model for detecting each input sequence to achieve the highest accuracy; 
(ii) \textit{quantity}: the multivariate dataset also consists a larger number of training samples compared to the univariate dataset, which helps to learn a better policy.

The kNN-single and kNN-sequence \cite{Taylor_SIGPLAN2018} schemes can achieve average delays that are lower than our proposed scheme. However, their average accuracy and F1-score are not competitive, as they are even worse than the {\it Edge} scheme.
Therefore, while their equivalent reward values (calculated using the reward function \eqref{eq:rewardFunction}) exceed those of the baseline schemes, they are lower than our proposed scheme---particularly in the multivariate case.
We also observe that {\it kNN-single} does not perform as well as {\it kNN-sequence} (which consists of three consecutive kNN-classifiers) in terms of accuracy, F1-score, and reward.

The {\it Adapted-BlockDrop} scheme \cite{Blockdrop_WuCVPR2018} achieves higher accuracy and F1-score than the IoT Device, Edge, Successive, kNN-single, and kNN-sequence schemes. However, it is outperformed by our proposed scheme in all the performance metrics (accuracy, F1-score, and delay).
In particular, with the univariate data, its average detection delay (301.91\,ms) is 4 times larger than that of our proposed scheme (76.12\,ms).
\begin{table}[t]
    \caption{Comparison among model selection schemes.}
    \label{tab:ExperimentResultDynamicScheme}
    \centering
    \begin{tabular}{l@{\hspace{0.8em}} l@{\hspace{0.8em}} c@{\hspace{0.8em}}   c@{\hspace{0.8em}}  c@{\hspace{0.8em}}  l@{\hspace{0.8em}} }
        \toprule
        \textbf{Dataset} & \textbf{Scheme}  &  \textbf{F1} & \textbf{Accuracy(\%)} & \textbf{Delay(ms)} & \textbf{Reward}\\%
        \midrule%
        \multirow{7}{*}{\STAB{\rotatebox[origin=c]{90}{Univariate}} } & \textbf{IoT Device}    & 0.465 & 93.68 & 12.4 & 48.39 \\%
        & \textbf{Edge}   & 0.800 & 98.63 & 257.43 & 45.36 \\%
        & \textbf{Cloud}$^{(b)}$  & 0.909 & 99.46 & 504.50 & 41.24\\%
        & \textbf{Successive} & 0.769 &  98.35 & 105.27 &   N/A   \\%
        & \textbf{kNN-single} & 0.588 & 96.15 & 31.29 & 49.27 \\%
        & \textbf{kNN-sequence} & 0.741 & 98.07 & 54.92 & 49.77 \\%
        & \textbf{Adapted-BlockDrop}$^{(a)}$ & 0.842 & 99.09 & 301.91 & 31.44\footnotesize{$^{(c)}$}\\%
        & \textbf{Proposed$^{(a)}$}  & {\bf 0.870}   & {\bf 99.11} & 76.12 & {\bf 49.82} \\%
        \midrule
        \multirow{7}{*}{\STAB{\rotatebox[origin=c]{90}{Multivariate}} } &\textbf{IoT Device}    & 0.848 & 93.19 & 591.0 & 351.18\\
        & \textbf{Edge}   & 0.951 & 97.59 & 667.30 & 362.16\\
        & \textbf{Cloud}$^{(b)}$  & 0.980 & 99.00 & 732.30 & 360.26\\ 
        & \textbf{Successive} & 0.911 &  95.79  & 626.16 &   N/A  \\%
        & \textbf{kNN-single} & 0.925  & 96.39 & 597.71 & 366.22  \\%
        & \textbf{kNN-sequence} & 0.929  & 96.59 & 598.79 & 367.07  \\%
        & \textbf{Adapted-BlockDrop}$^{(a)}$ & 0.962 & 98.13 & 657.31 & 420.17\footnotesize{$^{(c)}$} \\%
        & \textbf{Proposed$^{(a)}$ }  & {\bf 0.984}   & {\bf 99.01} & 654.74  & {\bf 371.61}\\
        \bottomrule
    \end{tabular}

    \vspace{0.3 mm}
    \footnotesize{$^a$ Average results from at least 3 trained policy networks ($\alpha=0.0005$ for the Proposed scheme). }\\%
    \footnotesize{$^{(b)}$ Cloud's marginal F1 and accuracy advantage are at the cost of a much higher delay.}\\%
    \footnotesize{$^{(c)}$ This reward value is calculated according to \cite{Blockdrop_WuCVPR2018} which results in a different range from the other values calculated by our reward function. So, these marked reward values are not comparable to the other reward values.}

\end{table}


In summary, and as also indicated in the last column ``Reward'', which is a convex combination of both accuracy and delay, our proposed adaptive scheme strikes the best tradeoff between accuracy and detection delay for univariate data with handcrafted feature representations of contextual information,
and achieves the best performance for these two metrics for multivariate data with encoded feature representations.
This is achieved by leveraging the distributed HEC architecture and our policy network that automatically selects the best layer to execute the detection task.

\begin{figure}[tb]
    \centering
    \subfloat[Univariate data]{
        \includegraphics[width=0.85\linewidth]{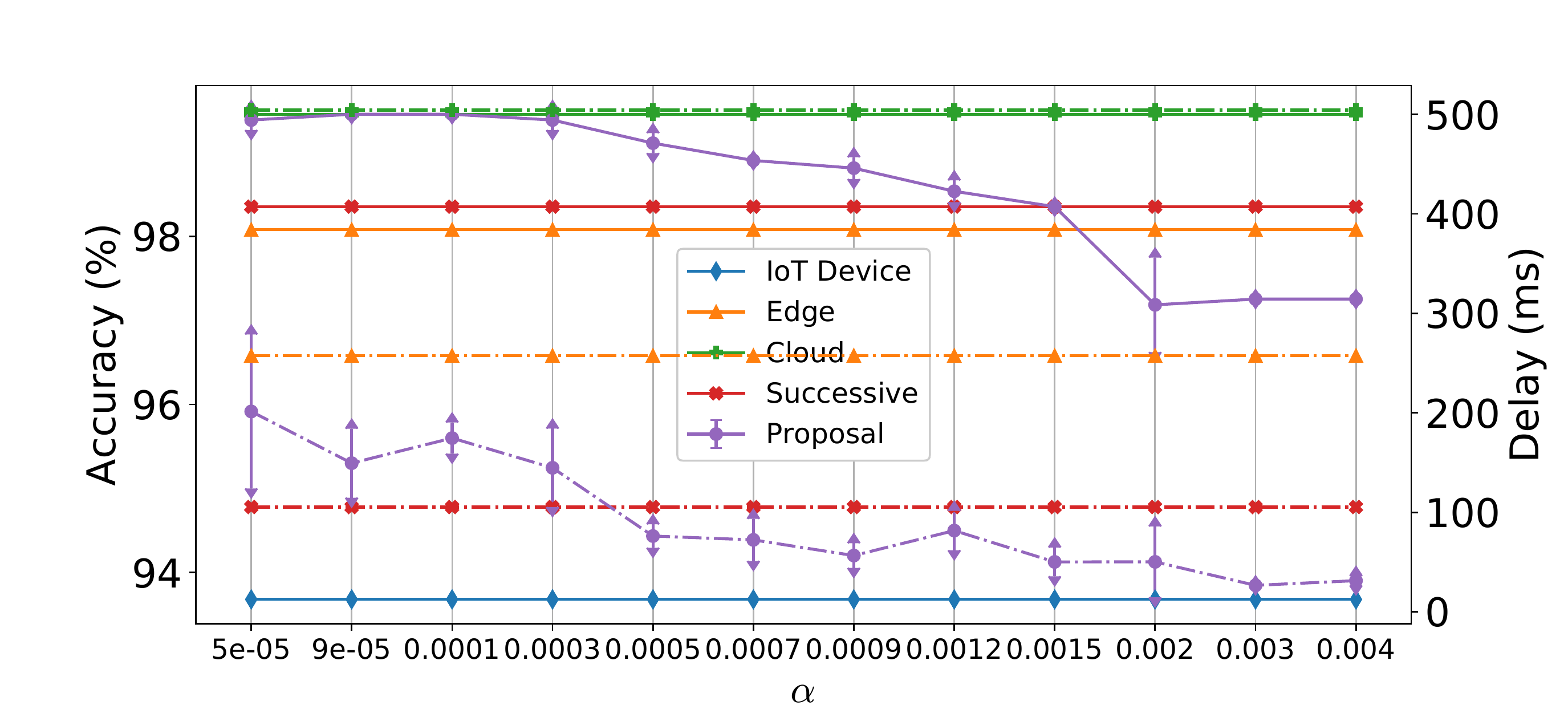}%
        \label{fig:tradeoffAccuracyDelayPowerDemand}
    }\quad
    \subfloat[Multivariate data]{
        \includegraphics[width=0.85\linewidth]{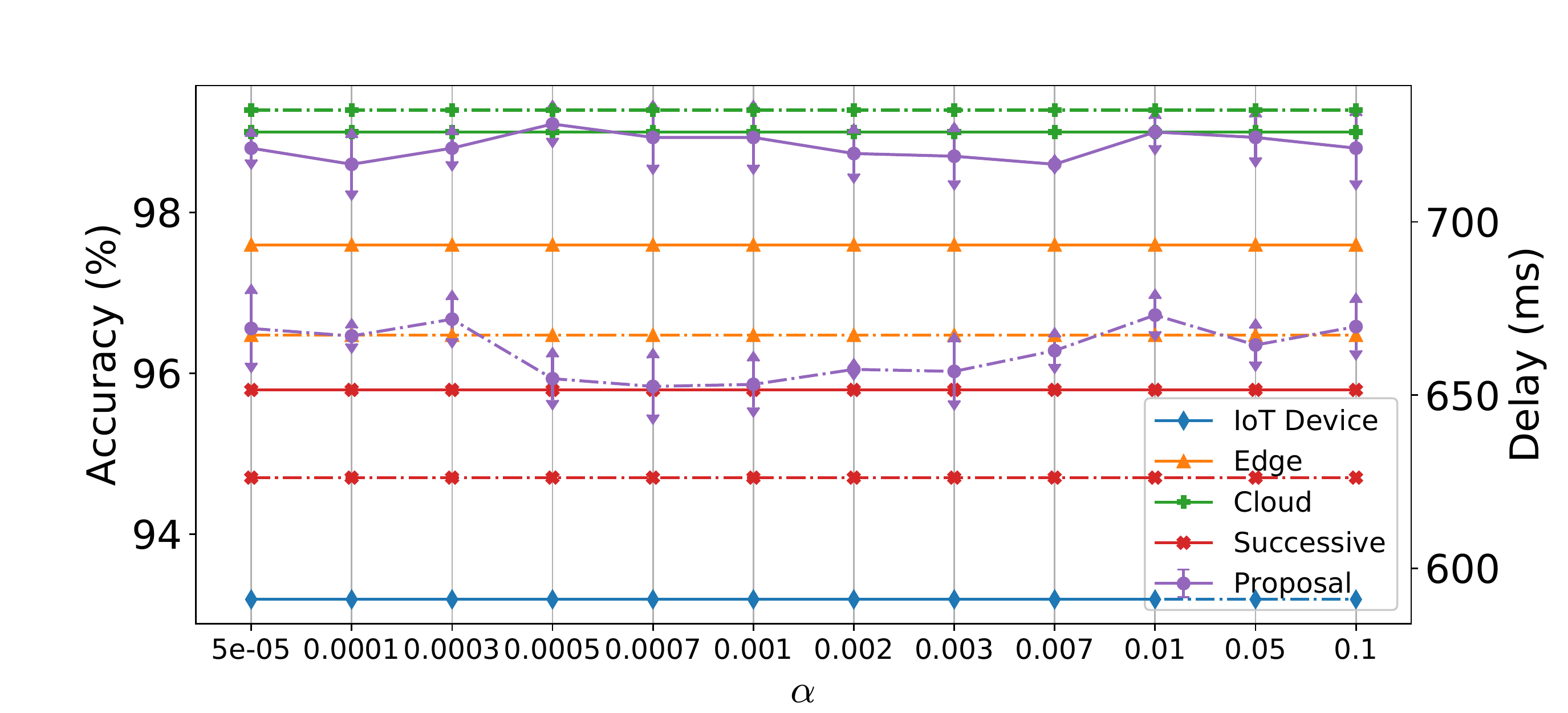}%
        \label{fig:tradeoffAccuracyDelayMhealth}
    }
    \caption{Accuracy (solid line) and delay (dashed line) of all evaluated schemes vs. different values of $\alpha$.}
     \label{fig:tradeoffAccuracyDelay}
\end{figure}

\textbf{Execution time of policy network:}
We measure the average execution time of the policy network on the actual IoT devices as follows: 8.4\,ms and 44.9\,ms for univariate and multivariate data, respectively.
For univariate data, although the execution time of our policy network (which is small) is comparable to the detection delay at the IoT layer (12.4\,ms), it is only 11\% of the overall average detection delay of the adaptive scheme (76.12\,ms).
For multivariate data, the execution time of the policy network is about 6.8\% of the average end-to-end detection delay of the adaptive scheme (654.74\,ms).
Overall, we can see that the execution time of the policy network as compared to the whole detection delay is negligible.

Compared to the state-of-the-art schemes, the policy network of the Adapted-BlockDrop scheme \cite{Blockdrop_WuCVPR2018} uses the same neural network architecture as our policy network; hence, we do not compare the execution time with the Adapted-BlockDrop scheme.
We measure the average execution time of the decision-making module of the kNN-single scheme on the IoT devices (Raspberry Pi3), and the results are 138.2\,ms and 159.4\,ms for univariate and multivariate data, respectively.
For the kNN-sequence decision making module with a series of three kNN classifiers, the execution time measured on the IoT devices are 118.3\,ms and 180.5\,ms for univariate and multivarate data, respectively.
Interestingly, kNN-single has longer execution delay than kNN-sequence for univariate data. We examined this observation and found that most of the selected actions of kNN-sequence are the IoT layer (i.e., exit after executing only the first kNN classifier). In addition, all kNN classifiers in kNN-sequence are binary classifiers whereas kNN-single is a multi-class classifier.
This explains the above counter-intuitive result.

The key observation is that, even though we have built kNN classifiers for embedded devices as lightweight as in \cite{Taylor_SIGPLAN2018} for comparison with our proposed method, the execution time of kNN-single and kNN-sequence schemes (118.3--180.5\,ms) is still much longer than that of our policy network (8.4--44.9\,ms), for both univariate and multivariate data.

\subsection{Cost Function: a Trade-off Between Accuracy vs Delay}
\label{subsec:Alpha}
We train different policy networks with different cost functions of which the tunable parameter $\alpha\in [0.00005, 0.004]$ for univariate data, and $\alpha \in [0.00005, 0.1]$ for multivariate data.
In Fig. \ref{fig:tradeoffAccuracyDelay}, we plot the mean and standard deviation of accuracy (solid line) and delay (dashed line) of the proposed scheme (over at least 3 trained policy networks for each $\alpha$), and the other baseline schemes (which are independent of $\alpha$).

For univariate data, we can see in Fig.~\ref{fig:tradeoffAccuracyDelayPowerDemand} that the accuracy, and detection delay of the proposed scheme gradually decreases when increasing $\alpha$.
With $\alpha \leq 0.0003$, the proposed scheme can achieve accuracy as high as the Cloud scheme, while the average delay of the proposed scheme significantly reduces by 60\%-90\% compared to that of the Cloud scheme.
Based on Fig.~\ref{fig:tradeoffAccuracyDelayPowerDemand}, we choose $\alpha=0.0005$ to get the best tradeoff between accuracy and delay, in which the accuracy only drops by 0.29\% compared to the Cloud scheme while the delay is even lower than that of the Successive scheme.

For multivariate data, we can see that the proposed scheme consistently achieves an accuracy as high as the Cloud scheme regardless of different $\alpha$.
As explained in Section~\ref{subsec:ComparisonModelSelectionScheme}, in some cases (e.g., $\alpha=0.0005$) the accuracy actually exceeds that of the Cloud scheme.
The average detection delay of the proposed scheme fluctuates around that of the Edge scheme.


\subsection{Accelerated Training of Policy Network}
\label{subsec:ResultAcceleration}
Comparison between average breakdown delays of training policy networks for each episode over the traditional training approach--\textit{sequential} and the accelerated training approach--\textit{parallel} is shown in Table~\ref{tab:ExperimentResultAccelerateTrainingPolicy}.
Similar to the sequential approach, under the new parallel approach as described in Section \ref{subsec:AcclerationTrainingPolicyNetwork}, we train the policy networks (with the same $\alpha=0.0005$ and number of episodes) for univariate and multivariate data to leverage parallel inference in distributed environments---distributed AD models at multiple HEC layers.
We can see in Table~\ref{tab:ExperimentResultAccelerateTrainingPolicy} that the average transmission delay on each training episode under the parallel approach is significantly reduced by 80.3\% and 82.9\% compared to that under the sequential approach for univariate and multivariate data, respectively.
For computing delay, we observe a slightly increasing computing delay under the parallel approach for univariate data, but a significantly decreasing computing delay with the parallel approach for multivariate data.
The delay reduction of the parallel approach in the multivariate data is because for edge and cloud models we use \texttt{CuDNNLSTM} units that leverage accelerated GPU hardware for parallel inference; while for the univariate data, the autoencoder AD models running inference on CPU device take longer time to compute a batch of multiple samples.
For training time, there is no difference between these two approaches.
Note that the training time of multivariate data is 10 times that of the univariate data because we split the large multivariate dataset into 10 mini-batches while the small univariate dataset is trained with one batch in each episode.
A combination of reduction delays of transmission and computing is signified when we account for the whole training process which consists of hundred episodes.
For example, for univariate data, the training policy network with 300 episodes took 7.6 hours under the sequential approach, but only 1.7 hours under the parallel approach.

\begin{table}[t]
    \caption{Comparison of sequential and parallel approaches in terms of average delay (broken down into three components).}
    \label{tab:ExperimentResultAccelerateTrainingPolicy}
    \begin{tabular}{ c@{\hspace{0.5em}}  c@{\hspace{0.5em}}  c@{\hspace{0.5em}}  c@{\hspace{0.5em}}  c@{\hspace{0.5em}}  }
        \toprule
	    \textbf{Dataset} & \textbf{Approach} &\textbf{Transmission delay(s)} & \textbf{Computing delay(s)} & \textbf{Training time(s)}  \\
         \midrule
         \textbf{Uni-} & {Sequential} & 48.834$\pm$5.696 &  0.661$\pm$0.188 & 0.013$\pm$0.006\\%
          \textbf{variate}$^a$ & {Parallel} & \textbf{9.605}$\pm$\textbf{2.290} & 0.958$\pm$0.697  & 0.015$\pm$0.004\\  %
          \midrule
         \textbf{Multi-} & {Sequential} & 420.78$\pm$53.69  &  351.30$\pm$43.67 & 0.141$\pm$0.01 \\
          \textbf{variate}$^b$ & {Parallel} & \textbf{72.17}$\pm$\textbf{9.66}& \textbf{78.06$\pm$13.14} & 0.15$\pm$0.12 \\
        \bottomrule
    \end{tabular}

\vspace{0.3 mm}
\footnotesize{$^a$ Training 300 episodes with 1 batch in each episode.}\\
\footnotesize{$^b$ Training 100 episodes with 10 mini-batches in each episode.}
\end{table}

\subsection{Contextual Information: Handcrafted vs Encoded Features}
\label{subsec:HandcraftedVsEncoded}

\begin{figure}[tb]
    \centering
    \includegraphics[width=0.95\linewidth]{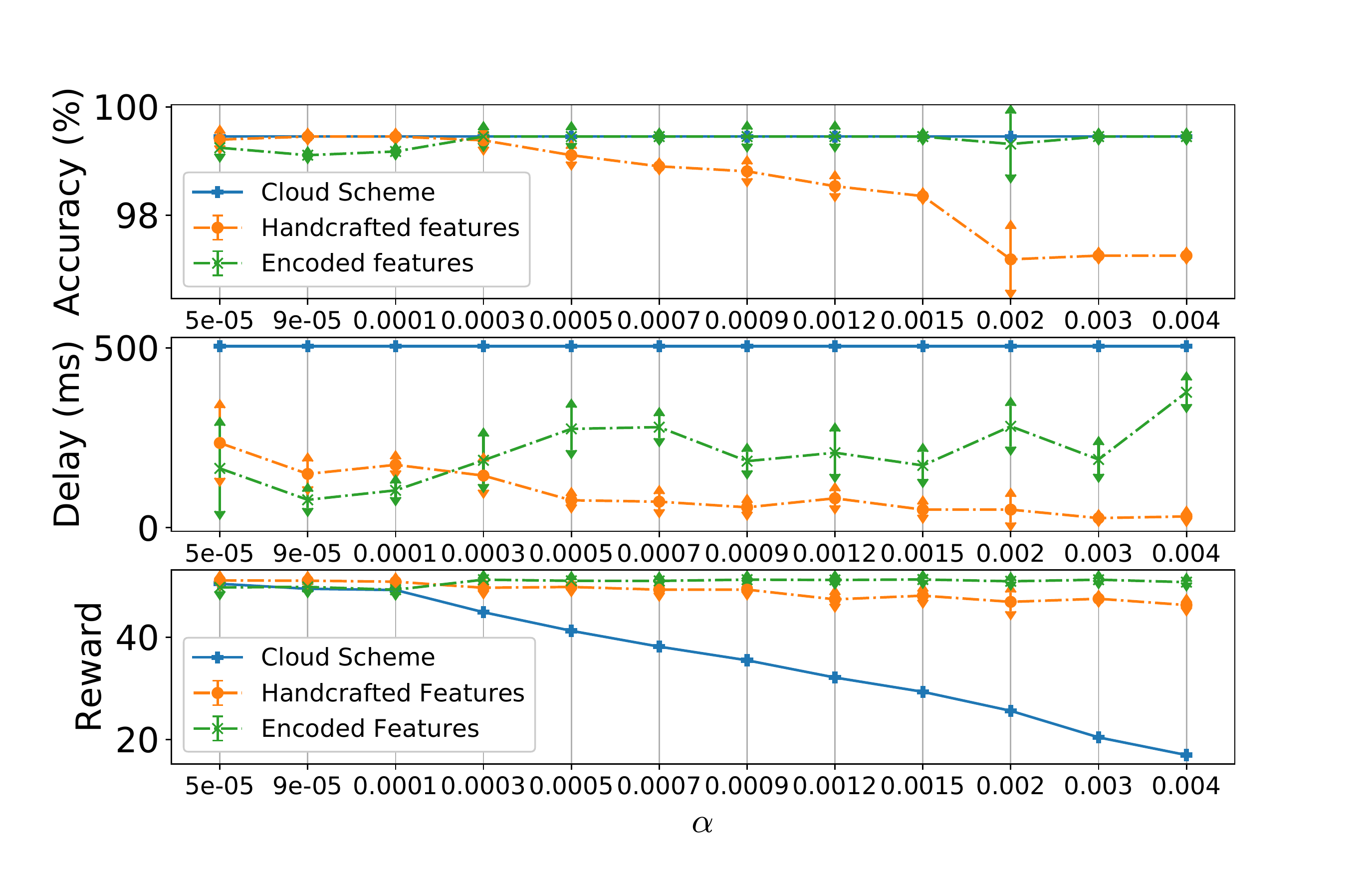}%
    \caption{Comparison between handcrafted features and encoded features for univariate data in terms of accuracy, delay, and reward under different $\alpha$.}
    \label{fig:CompareFeatureEngineeringEncodedVector}
\end{figure}

In Section~\ref{subsec:ComparisonModelSelectionScheme}, we have seen that the encoded states from LSTM-encoder for multivariate data, which capture good representations of contextual information, help the policy network consistently outperform the Cloud scheme in terms of accuracy.
In this section, we verify this claim for univariate data.
Instead of using handcrafted engineering features as the contextual information for univariate case, we use an encoded vector (201 dimensions) from the encoder model as the contextual state for training the policy network.
We still use a single hidden layer neural network with 500 hidden units as a policy network, then train multiple policy networks with different $\alpha$ under this new setup.
Fig.~\ref{fig:CompareFeatureEngineeringEncodedVector} shows a comparison between the handcrafted engineering features and the encoded representations of input in terms of accuracy, delay, and reward.
We can see that the encoded representations boost the performance of the policy network, which allow it to achieve accuracy nearly as high as the Cloud scheme while detection delay is significantly lower than that of the Cloud scheme.
This result is consistent with what we found for multivariate data.

However, we note that the detection delay under the new setup is higher than it is for handcrafted features.
Therefore, if one needs to balance between accuracy and detection delay, good handcrafted engineering features are a good option.
But it requires a domain-specific knowledge and lots of effort to find good feature representations.

\section{Conclusions}
\label{sec:conclusions}

We identify three issues in existing IoT anomaly detection approaches, namely using one universal model to fit all data, lopsided focus on accuracy, and lack of local analysis.
We then propose an adaptive approach to anomaly detection for both univariate and multivariate IoT data in distributed HEC.
It constructs multiple distributed AD models based on an autoencoder and LSTM with increasing complexity, and associates each of them with an HEC layer from bottom to top, i.e., IoT devices, edge servers, and cloud.
Then, it uses a reinforcement learning-based adaptive scheme to select the best-suited model on the fly based on the contextual information of input data.
The scheme consists of a policy network as the solution to a contextual-bandit problem, characterized by a single-step MDP.
We also presented the accelerated method for training the policy network to take advantage of the distributed AD models of the HEC system.
We implemented the proposed scheme and conducted experiments using two real-world IoT datasets on the HEC testbed that we built.
By comparing with other baseline and some state-of-the-art schemes, we show that our proposed scheme strikes the best accuracy-delay tradeoff with the univariate dataset, and achieves the best accuracy and F1-score while the delay is negligibly larger than the best scheme with the multivariate dataset.
For example, in the univariate case, the proposed scheme reduces detection delay by 84.9\% while retaining comparable accuracy and F1-score, as compared to the cloud offloading approach.





\bibliographystyle{ACM-Reference-Format}
\bibliography{references_acm}


\end{document}